\documentclass[10pt,twocolumn,letterpaper]{article}

\usepackage{multirow}

\usepackage{cvpr} 

\usepackage{bm}

\definecolor{cvprblue}{rgb}{0.21,0.49,0.74}
\usepackage[pagebackref,breaklinks,colorlinks,allcolors=cvprblue]{hyperref}
\usepackage[table,xcdraw]{xcolor}

\def\paperID{*****} 
\def\confName{CVPR}
\def\confYear{2026}

\title{Harmonized Feature Conditioning and Frequency-Prompt Personalization for Multi-Rater Medical Segmentation}

\author{
Sanaz Karimijafarbigloo$^{1}$ \quad
Armin Khosravi$^{2}$ \quad
Alireza Kheyrkhah$^{3}$\\
Reza Azad$^{1}$ \quad
Mauricio Reyes$^{4,5}$ \quad
Dorit Merhof$^{1,6}$\\[0.5em]
$^{1}$Faculty of Informatics and Data Science, University of Regensburg, Regensburg, Germany\\
$^{2}$Sharif University of Technology, Tehran, Iran\\
$^{3}$Iran University of Science and Technology, Tehran, Iran\\
$^{4}$ARTORG Center for Biomedical Research, University of Bern, Bern, Switzerland\\
$^{5}$Department of Radiation Oncology, University Hospital Bern, University of Bern, Bern, Switzerland\\
$^{6}$Fraunhofer Institute for Digital Medicine MEVIS, Bremen, Germany
}
\begin{document}
\maketitle
\begin{abstract}
Multi-rater medical image segmentation captures the inherent ambiguity of clinical interpretation, where diagnostic boundaries vary across experts and imaging devices. Existing approaches often reduce this diversity to consensus labels or treat rater differences as noise, resulting in overconfident and poorly calibrated models. We propose a harmonized probabilistic framework that disentangles acquisition artifacts from genuine annotator variability through adaptive feature conditioning and frequency-domain personalization. A lightweight Harmonizer Network implicitly models scanner-specific artifacts and performs dynamic feature modulation to standardize latent representations, ensuring that uncertainty reflects anatomy rather than noise. To represent rater-specific styles, we introduce a novel High-Frequency Prompt Modules that operate in the spectral domain to encode annotator-dependent boundary precision and textural sensitivity. These prompts adaptively modulate harmonized features to produce personalized yet anatomically consistent segmentations. Furthermore, a Generalized Energy Distance (GED)–based regularization aligns the generative distribution with empirical annotation variability, promoting diversity where experts disagree and consensus where they converge. Experiments on LIDC-IDRI and NPC-170 show SOTA aggregated and individualized segmentation, with notable GED reductions and improved Dice scores, especially on noisy cases. Beyond accuracy, the model exhibits clinically meaningful uncertainty. Confidence rises in agreement regions and declines in ambiguous areas, supporting its use as a reliable and interpretable tool for multi-expert clinical workflows. \href{https://github.com/sanazkarimi/harmonizer}{GitHub code}

\end{abstract}    
\section{Introduction}
\label{sec:intro}

Multi-rater medical image segmentation utilizes annotations from multiple clinical experts to delineate anatomical structures, acknowledging the inherent ambiguity and inter-observer variability in medical imaging \cite{li2024qubiq}. This variability renders the notion of a single, definitive ground truth untenable and stems from multiple factors: the subtlety and ambiguity of imaging features, differences in annotators’ training and clinical judgment, and the intrinsic uncertainty associated with early-stage or borderline pathologies \cite{wang2025learning}. Inter-rater variability is a significant challenge in medical image segmentation, particularly when delineating structures such as lung nodules, prostate tumors, or retinal layers, where interpretations often diverge among specialists (e.g., radiologists, oncologists, or ophthalmologists) \cite{kumari2025annotation}.
Traditional preprocessing strategies, such as majority voting or annotation averaging, are ill-suited to capture the nuances of interobserver variability in medical imaging. These aggregation-driven approaches discard critical information about annotator disagreement and underlying uncertainty, often biasing models toward overly confident or misleading labels \cite{cheplygina2019notsosupervised}. Recognizing this limitation, multi-rater medical image segmentation has emerged as a paradigm that explicitly leverages, rather than suppresses, expert disagreement. In contrast to conventional supervised learning, these methods utilize multiple annotations to infer probabilistic label fusion, model annotator-specific behavior, or generate a distribution of plausible segmentations \cite{kohl2018probabilistic}, thereby better reflecting the inherent uncertainty in clinical data and enhancing model interpretability, robustness, and clinical trustworthiness \cite{ye2023confidence,baumgartner2019phiseg}. STAPLE \cite{warfield2004simultaneous}, an early statistical approach, estimated a latent ground truth but relied on restrictive parametric assumptions. The advent of deep learning has since enabled a shift toward expressive probabilistic frameworks: the Probabilistic U-Net \cite{kohl2019hierarchical} introduced sample-level uncertainty modeling via latent distributions, paving the way for subsequent advances such as expert-specific disentanglement (D-LEMA \cite{mirikharaji2021d}), relational annotator modeling \cite{ji2021learning}, and prompt-guided personalization (D-Persona \cite{wu2024diversified}).

Despite these advances, most existing multi-rater segmentation frameworks still operate primarily in the spatial domain, making them vulnerable to scanner-specific noise, acquisition artifacts, and inconsistent annotation quality. Such perturbations can propagate through the latent space, conflating clinically meaningful uncertainty with irrelevant variability and hindering cross-scanner generalization. Previous efforts like D-Persona~\cite{wu2024diversified} and DiffOSeg~\cite{Zhang2025DiffOSeg} introduced annotator-specific latent codes for personalized predictions, yet they remain limited to feature-level modulation and do not explicitly disentangle noise effects from diagnostic variability. Moreover, recent analyses of latent probabilistic models~\cite{valiuddin2024investigating} show that naive latent distributions often become sparse or under-regularized, reducing their ability to represent structured variation across scanners. To address these issues, we propose a lightweight Harmonizer Network that adaptively conditions features across scales. Hence, it learns data-driven feature modulation to adapt to scanner variations, reducing noise and yielding a latent space that captures true anatomy.


Building on this harmonized foundation, our innovative framework integrates personalization directly within the frequency domain, where stylistic differences among annotators, such as boundary precision, texture sensitivity, and lesion delineation style, manifest as high-frequency cues. To capture these variations, we introduce High-Frequency Prompt Modules that decompose features via wavelet transforms and adaptively modulate the harmonized latent representation to reflect each expert’s diagnostic tendencies. This spectral personalization allows the model to synthesize expert-specific segmentations without retraining or replicating the backbone, maintaining both efficiency and interpretability. In parallel, a GED-based regularization aligns the learned predictive distribution with empirical annotation variability, reinforcing consensus where experts agree and promoting diversity where they diverge. 
Together, these components form a unified probabilistic framework that disentangles scanner-induced artifacts from genuine annotator variability, providing a lightweight yet effective solution for harmonized and personalized multi-rater segmentation.

\section{Related work}

\subsection{Label Fusion}
Label fusion techniques in multi-rater medical image segmentation aggregate diverse expert annotations into a single consensus ground truth, often referred to as meta-segmentation, to mitigate inter-observer variability and provide a reliable reference for training deterministic models. Early methods, like majority voting or STAPLE~\cite{warfield2004simultaneous}, probabilistically weight annotations based on estimated rater performance to fuse labels, assuming an underlying true segmentation exists but is obscured by individual biases.

Deep learning has introduced more advanced strategies for label fusion in multi-rater medical image segmentation. Soft Label Fusion~\cite{lourencco2021using} preserves inter-rater variability by averaging annotations into probabilistic soft labels, reducing overconfidence near uncertain boundaries and enhancing calibration relative to deterministic label fusion methods. D-LEMA~\cite{mirikharaji2021d} advances this by using voxel-wise reliability maps to adaptively weight annotations across spatial dimensions, enabling more robust consensus in regions of local disagreement. Subsequent methods incorporate task-specific context; for instance, DiFF~\cite{wu2024calibrate} presents a diagnosis-aware framework that weights annotator labels based on their relevance to diagnostic outcomes, jointly modeling segmentation and disease classification to produce a fused representation that is both consistent and clinically pertinent. Further developments explicitly model uncertainty. DiffOSeg~\cite{Zhang2025DiffOSeg}, a two-stage diffusion-based architecture, establishes a robust population-level fusion followed by the generation of personalized expert-style segmentations using adaptive prompts. 
This integration of agreement-centric and preference-oriented learning reflects the evolution from simple static averaging to dynamic, expert-informed representations.

Despite their utility, label fusion methods often overlook the inherent diversity of expert opinions, risking an over-aggregated representation that discards valuable uncertainty. They also assume a definitive meta-segmentation, which may fail in ambiguous cases where consensus doesn’t equal truth. These methods relate to diversity-preserving approaches by supplying fused labels for generative training, yet they emphasize a single reliable outcome over multiple plausible hypotheses.

\subsection{Diversity-preserving}
Diversity-preserving methods model the conditional distribution of plausible labels given an input image to generate multiple diverse yet realistic segmentation samples, capturing annotation ambiguities without collapsing to a single consensus. Kohl et al.~\cite{kohl2018probabilistic} pioneered the Probabilistic U-Net, a one-stage CVAE-integrated framework with a low-dimensional latent space where variants are encoded; it uses a prior network for image-conditioned distribution estimation and a training-only posterior for learning, producing unlimited samples via stochastic sampling. However, its isotropic Gaussian latent distribution hinders modeling of complex, multimodal annotation variations.

Subsequent works enhance hierarchical structures to propagate stochasticity across resolutions for improved sample quality. Baumgartner et al.~\cite{baumgartner2019phiseg} propose PHiSeg, a one-stage hierarchical CVAE with separate latent variables per resolution level in a U-Net-like architecture, ensuring stochasticity through all skips; it generates unlimited samples matching ground-truth distributions on prostate MRI and LIDC, outperforming Probabilistic U-Net in GED and NCC metrics. Similarly, the Hierarchical Probabilistic U-Net~\cite{monteiro2020stochastic} decomposes latents coarsely to finely, modeling multi-scale ambiguities for instance segmentation tasks like neuronal cells; it achieved superior GED and HM-IoU on the LIDC dataset compared to PHiSeg. 
Diffusion-based models advance diversity through stochastic denoising processes. CIMD et al.~\cite{rahman2023ambiguous}, a one-stage diffusion framework that models ambiguity through latent space diffusion without additional priors, produces diverse masks across CT, ultrasound, and MRI modalities; it introduces the CI score and outperforms VAE baselines in D\_max and collective insight. 
Although these approaches implicitly model diverse annotation styles across multiple raters, they fail to capture true personalized segmentation behavior tailored to individual experts, remaining centered on consensus-level variability rather than expert-specific adaptation.

\subsection{Personalized adaptation}

Personalized and expert-aware modeling approaches aim to produce segmentation predictions tailored to the distinct annotation styles and preferences of individual experts. Early methods such as Who Said~\cite{guan2018said} and Disentangle~\cite{zhang2020disentangling} modeled annotator-specific biases using confusion matrices and probabilistic noise models, disentangling true anatomical structures from annotation noise to enable robust consensus estimation while retaining individual annotator characteristics. Based on probabilistic frameworks, BayesianNN~\cite{hu2023inter} introduced a shared encoder with rater-specific Bayesian decoders that feature dedicated attention mechanisms. This enabled efficient modeling of annotator-specific styles, outperforming classical probabilistic baselines such as PHiSeg.

Personalization evolved toward structured, interpretable designs. TAX~\cite{cheng2023tax} enabled transparency via annotator-specific kernels and a prototype bank, while That Label’s Got Style~\cite{Zepf2023That} disentangled style from uncertainty using one-hot style conditioning. Prompt-based methods unified this direction: PU-Net~\cite{Wang2024MultiRaterPrompt} introduced rater-specific and aggregated prompts, treating annotators as domains; D-Persona~\cite{wu2024diversified} enhanced diversity through a constrained latent space queried by expert prompts. Most recently, DiffOSeg~\cite{Zhang2025DiffOSeg} integrated agreement modeling and personalization in a diffusion framework, with probabilistic aggregation in Stage I and adaptive prompting in Stage II, replicating expert styles without architectural redundancy.

\subsection{Motivation}
Despite notable progress, existing multi-rater segmentation frameworks remain limited by their sensitivity to scanner-dependent noise and their inability to distinguish data-level ambiguity from genuine inter-rater variability. Scanner artifacts, acquisition heterogeneity, and inconsistent annotation quality often entangle within the latent space, obscuring clinically meaningful uncertainty and degrading generalization. Meanwhile, rater-specific nuances in boundary precision or texture perception demand adaptive modeling rather than static fusion. Our framework addresses these intertwined challenges through a unified design centered on the Harmonizer Network, which mitigates acquisition-induced artifacts while preserving annotator diversity via frequency-aware conditioning. This dual-level harmonization lets the model coherently represent both ambiguity sources, stabilizing features and capturing rater-specific behavior.

\begin{figure*}[!tbh]
    \centering
    \includegraphics[width=0.93\linewidth]{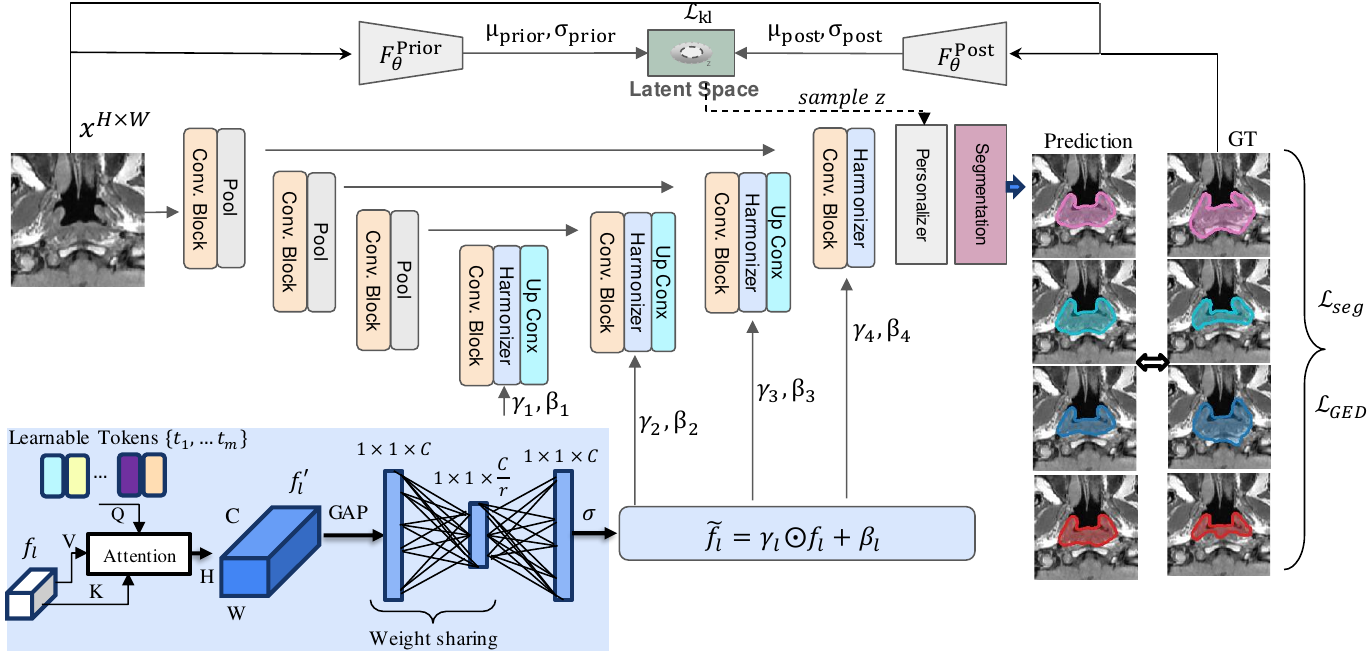}
    \caption{Illustration of the proposed Harmonizer Network for multi-rater medical image segmentation.}
    \label{fig:main}
\end{figure*}

\section{Method}
\label{sec:method}
As illustrated in \Cref{fig:main}, our unified probabilistic framework aims to learn the conditional distribution $p(y \mid x)$ of plausible segmentation masks given an input medical image $x$, while explicitly disentangling two intertwined sources of ambiguity: data-level noise, originating from scanners and acquisition heterogeneity, and rater-level variability, driven by subjective diagnostic interpretation. To address these challenges, we introduce two cooperating modules within a single architecture: a Noise Harmonizer that standardizes latent representations across acquisition domains, and a Personalization Module that adapts the harmonized features in the frequency domain to capture rater-specific nuances. Both operate under a shared probabilistic backbone, ensuring coherent uncertainty modeling across spatial and spectral dimensions.

We build our base model upon the probabilistic formulation of the Probabilistic U-Net~\cite{kohl2018probabilistic} approach, which models the conditional distribution of segmentation masks $y$ given an image $x$ through a low-dimensional latent variable~$z$:
\begin{equation}
    p_\theta(y \mid x) = \int p_\theta(y \mid x, z) \, p_\theta(z \mid x) \, \mathrm{d}z,
\end{equation}
where $p_\theta(z \mid x)$ is the learned prior and $p_\theta(y \mid x, z)$ is the decoder likelihood.

Given an image $x \in \mathbb{R}^{H \times W}$ and its $n$ corresponding rater annotations $\mathcal{A} = \{A^{(1)}, \dots, A^{(n)}\}$, the encoder network $F^{\text{enc}}_\theta$ extracts spatial features $f = F^{\text{enc}}_\theta(x)$, while the prior $F^{\text{prior}}_\theta$ and posterior $F^{\text{post}}_\theta$ networks predict Gaussian parameters $\{\mu_{\text{prior}}, \sigma_{\text{prior}}\}$ and $\{\mu_{\text{post}}, \sigma_{\text{post}}\}$, respectively. Latent codes $z$ are sampled via the reparameterization trick $z = \mu + \sigma \odot \epsilon$, $\epsilon \sim \mathcal{N}(0, I)$, and concatenated with feature maps $f$ before being fed to a decoder $F^{\text{dec}}_\theta$ to produce segmentation hypotheses $\hat{y}$ \cite{kohl2018probabilistic}.

\subsection{Noise Harmonizer}
The Noise Harmonizer $\mathcal{H}_\phi^{(n)}$ mitigates scanner- and acquisition-induced artifacts that distort latent features and confound the learning of true anatomical variability.  
Given decoder features $f_l$ at layer $l$, the harmonizer predicts spatially broadcasted modulation parameters $(\gamma_l, \beta_l)$ conditioned on the features and a learnable bank of artifact tokens $t=\{t_1,\dots,t_M\}$ representing characteristic noise patterns:
\begin{equation}
\begin{split}
[\gamma_l, \beta_l] &= \mathcal{H}_\phi^{(n)}(f_l, {t})
= W_2\bigl( \sigma\bigl( W_1 (GAP(f'_l)) \bigr) \bigr), \\
f'_l &= 
     \operatorname{Softmax}\left( \frac{Q_j K_j^\top}{\sqrt{D_h}} \right) V_j,
\end{split}
\end{equation}

\noindent where K and V are generated from $f_l$, while Q is obtained from learnable tokens. $\text{GAP}(\cdot)$ denotes global average pooling, and $\sigma$ denotes a non-linearity.  
The harmonized output:
\begin{equation}
\tilde f_l = \gamma_l \odot f_l + \beta_l ,
\end{equation}
with $\odot$ denoting element-wise multiplication.  
This affine transformation acts as a data-driven normalization mechanism that dynamically suppresses intensity drifts, motion artifacts, and domain biases without requiring explicit knowledge of noise distribution.  
Weight sharing across layers imposes a cross-scale regularization, ensuring consistent denoising behavior throughout the network.

The learned parameters $(\gamma_l,\beta_l)$ implicitly encode the acquisition condition, guiding the network produce anatomically stable latent codes $z$ that are invariant to scanner variability, yet expressive to capture structural uncertainty.

\subsection{Rater-Level Adaptation}
While the Noise Harmonizer standardizes data-level features, the Personalization Module $\mathcal{H}_\phi^{(p)}$ adapts the latent representation to reflect rater-specific tendencies in the frequency domain.  
This module assumes that subtle stylistic variations, such as differences in edge delineation, texture sensitivity, or lesion extent, manifest predominantly in the high-frequency spectrum.

Given feature map $X \in \mathbb{R}^{H \times W \times D}$, we begin by reducing its dimensionality through a linear projection $\widetilde{X} = XW_d$, where $W_d \in \mathbb{R}^{D \times \frac{D}{4}}$. The transformed feature $\widetilde{X}$ is then decomposed in the spectral domain using the Discrete Wavelet Transform (DWT) with a Haar basis, resulting in four sub-bands that separate low- and high-frequency information,
$\hat{X} = [X_{LL}, X_{LH}, X_{HL}, X_{HH}] \in \mathbb{R}^{\frac{H}{2} \times \frac{W}{2} \times \frac{D}{4}},$ where $X_{LL}$ represents the structural outline, while $X_{LH}$, $X_{HL}$, and $X_{HH}$ contain detailed texture and edge information that are often interpreted differently by individual raters. These high-frequency maps are concatenated into a unified tensor $X_H = [X_{LH}, X_{HL}, X_{HH}] \in \mathbb{R}^{\frac{H}{2} \times \frac{W}{2} \times {D'}}$, $(D' = \frac{3D}{4})$, which serves as input to the Rater-Aware Prompt Projection block.

\begin{figure}[t]
    \centering
    \includegraphics[width=1\linewidth]{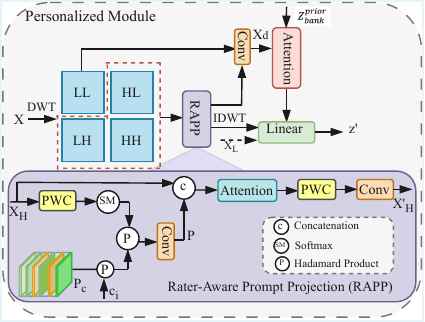}
    \caption{Proposed frequency recalibration method for personalized segmentation.}
    \label{fig:Personalized Module}
\end{figure}

\noindent\textbf{{Personalized Module:}} The personalized module (shown in \Cref{fig:Personalized Module}) is designed to encode annotator-specific preferences in the spectral domain and to propagate these latent representations into the decoding process. 
Within this block, 
\(N\) prompt components  \({P_c} \in \mathbb{R}^{N \times \frac{H}{2} \times \frac{W}{2} \times {D'}}\)
are used to represent potential annotation biases.
A learnable personalized weight $c_i$ modulates these components, enabling the model to generate person-specific tokens.
The model infers an adaptive weighting vector \(w \in \mathbb{R}^{D'}\) from $X_H$ to combine these codes into a context-conditioned prompt:
\begin{equation}\small
\label{eq:rater_prompt}
\begin{split}
    \mathbf{w} = \textbf{Softmax}(\textbf{PWC}(X_H)),\\ 
    {P} = \textbf{Conv}_{3\times3}\left(\sum_{c=1}^{{D'}} \mathbf{w} (c_i(P_c))\right).
\end{split}
\end{equation}

This prompt encapsulates rater preference information in the spectral domain, providing adaptive modulation over high-frequency features.

The prompt and spectral features interact through an attention mechanism that aligns image textures with the inferred rater biases, producing refined frequency responses:
{\small
\begin{equation}
\label{eq:rater_attention_block}
X'_H = \textbf{Conv}_{1\times1}(\textbf{Attention}(X_H, P)).
\end{equation}
}


For the attention mechanism, we employ the Large Kernel Attention module \cite{guo2023visual}. Subsequently, a $3{\times}3$ convolution is applied to the recalibrated high-frequency components $X'_{H}$ together with the low-frequency map $X_L$ to maintain local spatial coherence, producing a compact contextual feature representation denoted $X_d$.
Next, we construct a memory of latent priors by drawing $M_z$ samples from the fixed prior distribution $\mathcal{D}_{\text{prior}}$, forming a prior bank $\mathbf{Z}^{\text{prior}}_{\text{bank}} \in \mathbb{R}^{D \times 1 \times M_z}$.
The localized feature $X_d$ serves as the query in a cross-attention operation, while the elements of $\mathbf{Z}^{\text{prior}}_{\text{bank}}$ act as keys and values, allowing contextual cues to be aligned with the learned latent prototypes.
Subsequently, an inverse discrete wavelet transform (IDWT) is performed over the concatenation of the refined high-frequency feature $X'_{H}$ and the preserved low-frequency signal $X_L$, re-synthesizing a full-spectrum representation.
The resulting feature, fused with the cross-attention output, passes through a linear projection that produces a rater-adaptive latent vector $z'$ encoding the annotator’s structural bias.
This personalized latent code is finally integrated with the shared backbone features, yielding individualized yet anatomically consistent segmentations.

Through this design, the model benefits from enhanced sensitivity to subtle variations in texture and edge information, attributes that often underlie inter-rater disagreement. The jointly learned latent prompt codes therefore provide a compact representation of annotation bias, enabling the network to modulate high-frequency features according to rater-specific tendencies while maintaining anatomical fidelity and segmentation stability.

\noindent\textbf{Personalized Decoding: }Given $z'_i$, the decoder produces a personalized segmentation $\hat{y}_i = F^{\text{dec}}_\theta(f, z'_i, i)$ for rater $r_i$. During training, $\hat{y}_i$ is supervised by the corresponding rater annotation $A^{(i)}$, with the shared backbone and prior frozen:
\begin{equation}
    \mathcal{L}_{\text{Seg}} = \sum_{i=1}^n \mathbb{E}_{z \sim p_\theta(z \mid x)} \big[ \mathcal{L}_{\text{seg}}(\hat{y}_i, A^{(i)}) \big].
\end{equation}
Because $\mathcal{P}_i$ is lightweight, personalization can be achieved efficiently even with limited rater-specific data, making the framework suitable for semi-supervised or few-shot personalization scenarios.

\subsection{Diversity-Preserving}
To model the distributional nature of expert disagreement, we formalize segmentation as a matching problem between two conditional distributions: the model-induced distribution $\mathcal{P}(y\!\mid\!x)$ and the empirical annotation distribution $\mathcal{A}(y\!\mid\!x)$. An ideal probabilistic model should minimize the discrepancy between these distributions while preserving variability among its own predictions. Formally, this discrepancy can be expressed through the GED~\cite{bellemare2017cramer, kohl2018probabilistic}, which provides a symmetric, unbiased measure of distance between distributions. Using a segmentation metric $d(\cdot,\cdot)$ (we employ $d = 1-\mathrm{IoU}$), the empirical loss becomes: 
\begin{align}
\mathcal{L}_{\text{GED}} =\; &
\frac{2}{K N} \sum_{k=1}^K \sum_{i=1}^N d(P_k, A_i) \notag \\
& - \frac{2}{K(K-1)} \sum_{1 \leq k < k' \leq K} d(P_k, P_{k'}),
\end{align}

\noindent where $\{P_k\}$ are $K$ segmentation samples drawn from the model and $\{A_i\}$ are $N$ expert annotations. The first term aligns the model’s predictive distribution with the observed annotation manifold (fidelity), while the second enforces diversity across generated samples, preventing degeneracy toward a single consensus mask. This formulation regularizes the model to generate anatomically plausible, uncertainty-aware segmentations aligned with expert variability.

\subsection{Probabilistic Objective}
The overall objective combines probabilistic reconstruction loss, KL divergence regularization, a harmonization penalty, and a GED-based distribution alignment constraint:
\begin{equation}\small
\begin{aligned}
\mathcal{L}_{\text{total}} &=
\mathbb{E}_{z\sim q_\theta(z\mid x,A^{(k)})}
\!\big[\mathcal{L}_{\text{seg}}(F^{\text{dec}}_\theta(f', z'), A^{(k)})\big] \\
&\quad + \lambda_{\text{KL}}\,
\mathcal{L}_{\mathrm{KL}}\!\big(q_\theta(z\mid x,A^{(k)}) \,\|\, p_\theta(z\mid x)\big) \\
&\quad + \lambda_{\text{harm}}\!\sum_{l=1}^L (\|\gamma_l-1\|_2^2 + \|\beta_l\|_2^2)
+ \lambda_{\text{GED}}\mathcal{L}_{\text{GED}}(y,\mathcal{A}).
\end{aligned}
\label{eq:unified_loss}
\end{equation}
Here, $\mathcal{L}_{\text{seg}}$ combines Dice and cross-entropy losses, while $\mathcal{L}_{\text{GED}}$ enforces statistical alignment between predicted and empirical annotation distributions.

\section{Experiments and Results}
\subsection{Dataset}
\textbf{LIDC-IDRI}: We evaluate our method on the LIDC-IDRI dataset~\cite{armato2011lidc}, a standard benchmark for studying inter-observer variability in thoracic CT segmentation. The dataset provides diagnostic CT scans annotated by up to four radiologists through a two-stage process: an independent (blinded) segmentation of nodules $\geq3$ mm, followed by a consensus (unblinded) review. This process introduces substantial diversity in boundary delineations and nodule extent. Following common practice~\cite{zhang2020disentangling, wang2023medical}, we extract {1,609 axial slices} from {214 patients}, resample all scans to {0.5 mm isotropic resolution}, and crop them into {128$\times$128} nodule-centered patches. A {four-fold cross-validation} at the patient level is employed for fair evaluation.

\noindent \textbf{NPC-170:}
We further validate our framework on the NPC-170 dataset~\cite{wu2024diversified}, which contains multi-modal MRI scans of 170 nasopharyngeal carcinoma (NPC) patients annotated by {four experienced radiation oncologists}. Each subject includes T1-weighted, T2-weighted, and contrast-enhanced T1 (T1c) modalities, with independent delineations of the gross tumor volume (GTVp). The dataset is divided into {100/20/50} patients for training, validation, and testing, corresponding to {6,134}, {1,126}, and {3,058} annotated slices, respectively. We follow \cite{wang2025noisy} to ensure fair comparison and consistent evaluation across studies.

\subsection{Implementation Details and Training Strategy}
Our framework is built upon the probabilistic U-Net backbone~\cite{kohl2018probabilistic}, extended with the proposed Noise Harmonizer, Personalization Module, and the GED loss for distributional alignment. All default hyperparameter values, including learning rate schedules, optimizer configuration, and loss weighting, follow the original probabilistic U-Net implementation unless otherwise stated. 
To ensure stable latent representation learning and clear separation between data-level and rater-level variability, training is performed in two sequential phases, while maintaining a unified architecture.

\noindent\textbf{Phase 1 — Latent Space and Noise Harmonization.}  
In the first phase, the rater-specific frequency adaptation head is excluded, and the probabilistic backbone is trained jointly with the Noise Harmonizer. This phase focuses on learning artifact-invariant and anatomically consistent latent features before introducing rater-dependent modulation. Training is conducted for {100 epochs} using the Adam optimizer with an initial learning rate of {1e-4}, $\beta_1 = 0.9$, $\beta_2 = 0.999$, and weight decay of $1\times10^{-5}$. The latent space dimensionality is set to $D = 6$, and a memory buffer of size $M = 100$ is maintained to stabilize the conditional prior distribution.

\noindent\textbf{Phase 2 — Frequency-based Personalization.}  
After the first phase converges, backbone parameters (encoder, decoder, and Noise Harmonizer) are frozen, and only the Personalization Module is trained for {150 epochs}. This phase fine-tunes the frequency-domain adaptation to align spectral features with rater-specific annotation patterns while keeping the shared probabilistic representation fixed. A reduced learning rate of {5e-5} is used to ensure smooth adaptation.

\noindent\textbf{Implementation Details:}  
All experiments are conducted on a single NVIDIA GeForce RTX~3090 GPU. The full model contains 30.31\,M parameters (baseline 30.11\,M, $\mathcal{H}_\phi^{(n)}$ 0.14\,M  and $\mathcal{H}_\phi^{(p)}$ 0.07\,M parameters), including both harmonization and personalization modules, and requires around 0.42\,s per forward pass during inference. The network is implemented in PyTorch with mixed-precision training for efficiency, using a fixed seed in all experiments.

\subsection{Evaluation Metrics}
To assess performance in multi-rater segmentation, we use metrics capturing set-level diversity and expert-specific fidelity. We follow established evaluation practices from prior work on probabilistic and personalized segmentation \cite{kohl2018probabilistic, ji2021learning, wang2023medical}, and use the same metrics as those employed in recent personalized segmentation studies \cite{wu2024diversified, kumari2025annotation}.
To evaluate the distributional alignment between predicted and expert segmentation distributions, we use the GED \cite{bellemare2017cramer}, a statistical metric quantifying both sample quality and internal diversity. However, as GED ignores voxel-level agreement, we complement it with the soft-thresholded Dice score \cite{ji2021learning, wang2023medical}, measuring overlap between soft predictions and soft annotations across multiple thresholds. This enables a fine-grained view of voxel-wise calibration and improves robustness to specific operating points.

To assess whether the predicted set captures expert variability, we use two coverage metrics. $\mathrm{Dice}_{\max}$ measures upper-bound coverage by selecting the best prediction for each annotation, while $\mathrm{Dice}_{\text{match}}$ enforces stricter one-to-one matching via optimal bipartite assignment. Together, they distinguish true diversity from redundancy.
Finally, personalized agreement is measured by the Dice score between each expert annotation and its corresponding individualized prediction, denoted $\mathrm{Dice}_{A(i)}$. Averaging across experts gives $\mathrm{Dice}_{\text{mean}}$, reflecting how well the model captures annotator-specific styles rather than collapsing to a global consensus.

\begin{table*}[t]
\caption{Comparison of distribution fitting and sampling diversity performance on LIDC–IDRI and NPC–170 datasets.}
\centering
\resizebox{1.0\linewidth}{!}{
\begin{tabular}{c|cc|cc|cc|cc}
\toprule[1.5pt]
\multirow{2}{*}{Method} 
& \multicolumn{4}{c|}{\textbf{LIDC\textendash IDRI Dataset}} 
& \multicolumn{4}{c}{\textbf{NPC\textendash170 Dataset}} \\
\cline{2-9}
& $GED\downarrow$ & $Dice_{soft}\uparrow$ & $Dice_{max}\uparrow$ & $Dice_{match}\uparrow$
& $GED\downarrow$ & $Dice_{soft}\uparrow$ & $Dice_{max}\uparrow$ & $Dice_{match}\uparrow$ \\
\midrule

Prob. U-Net \cite{kohl2018probabilistic} (\#10) &0.2181 &88.79 &88.60 &88.43 &0.3585 &81.24 &\textbf{84.25} &80.18 \\
Prob. U-Net \cite{kohl2018probabilistic} (\#30) &0.2169 &88.79 &88.80 &88.73 &0.3536 &81.22 &84.19 &80.14 \\
Prob. U-Net \cite{kohl2018probabilistic} (\#50) &0.2168 &88.80 &88.87 &88.81 &0.3528 &81.19 &84.19 &80.13 \\
D-Persona \cite{wu2024diversified} (\#10) &0.1461 &90.24 &90.75 &90.51 &0.2314 &83.21 &83.03 &80.18 \\
D-Persona \cite{wu2024diversified}(\#30) &0.1375 &90.42 &91.23 &91.16 &0.2090 &83.74 &82.78 &80.98 \\
D-Persona \cite{wu2024diversified}(\#50) &{0.1358} &{90.45} &{91.37} &{91.33} &{0.1978} &{84.01} &{82.79} &{81.69} \\
\midrule
\rowcolor[HTML]{C8FFFD} 
Harmonizer Network (\#10) &0.1164 &91.54 &\textbf{92.28} &91.70 &0.1960 &84.27 &82.21 &81.18\\
\rowcolor[HTML]{C8FFFD} 
Harmonizer Network (\#30) &0.1063 &91.79 &92.27 &91.84 &0.1764 &84.82 &82.23 &82.36 \\
\rowcolor[HTML]{C8FFFD} 
Harmonizer Network (\#50) &\textbf{0.1048} &\textbf{91.81} &\textbf{92.28} &\textbf{91.94} &\textbf{0.1758} &\textbf{84.83} &{82.26} &\textbf{82.65} \\
\bottomrule[1.5pt]
\end{tabular}}
\label{tab:dual_dataset}
\end{table*}
\vspace{-2mm}
\begin{figure*}[th]
    \centering
    \includegraphics[width=1\linewidth]{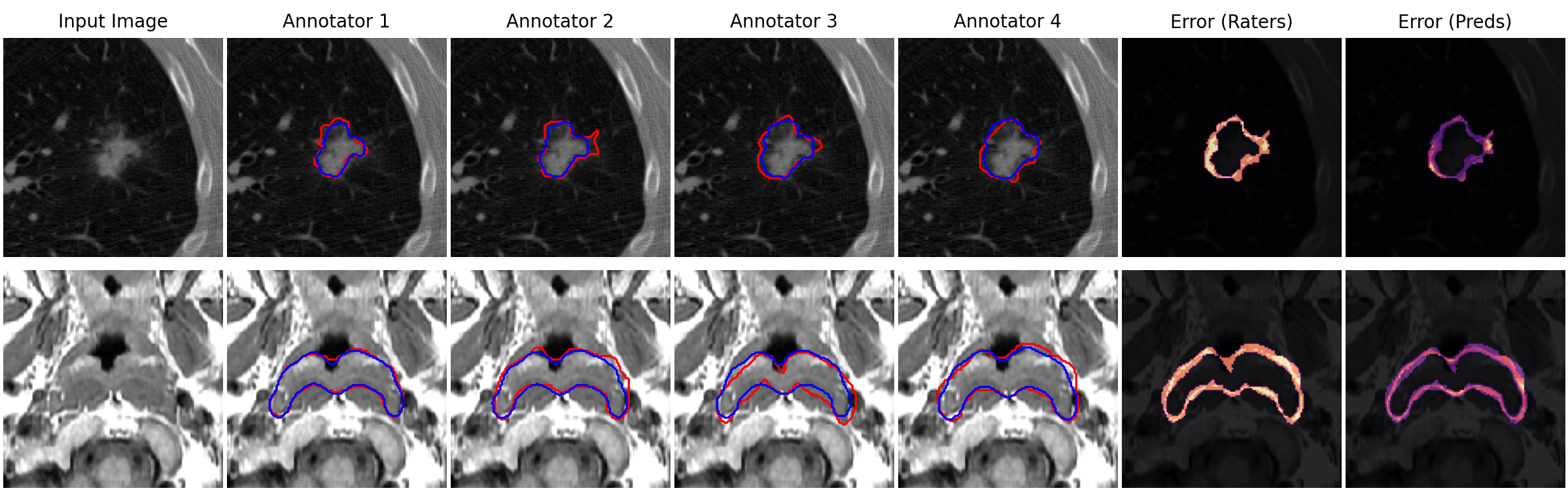}
    \caption{Visualization results on the LIDC (first row) and NPC-170 (second row) dataset with multi-rater annotations and the corresponding error map. The red boundary indicates the ground truth, while the blue boundary represents the predicted mask.}
    \label{fig:resmain}
\end{figure*}
\subsection{Results}
\label{sec:results}
\paragraph{Distributional fit and diversity of the sample set:}
As summarized in Table~\ref{tab:dual_dataset}, the Harmonizer prior in Phase~\uppercase\expandafter{\romannumeral1} achieves the strongest distributional alignment across both LIDC-IDRI and NPC-170 datasets. While the Probabilistic U-Net~\cite{kohl2018probabilistic} improves marginally with more samples $K$, its latent priors remain poorly calibrated, often generating redundant or anatomically inconsistent hypotheses. In contrast, our phase~\uppercase\expandafter{\romannumeral1} framework, equipped solely with the Noise Harmonizer and the GED-based distributional constraint, produces a latent manifold that is cleaner, more compact, and better aligned with the empirical annotation distribution. As $K$ increases from $10$ to $50$, GED steadily decreases and Dice$_{\text{soft}}$ improves, indicating that the model systematically expands its coverage of plausible annotations without over\textendash dispersing predictions. Compared with D-Persona~\cite{wu2024diversified}, which learns a spatially modulated latent prior without explicit noise disentanglement, our method first suppresses scanner- and acquisition- dependent artifacts before modeling inter-rater variability. This separation enables a more faithful mapping between anatomical uncertainty and sample diversity, yielding consistently lower GED (0.105 vs. 0.136 on LIDC and 0.175 vs. 0.197 on NPC-170) and higher Dice$_{\text{soft}}$. In essence, the Harmonizer prior learns where uncertainty matters,focusing diversity on clinically ambiguous boundaries while preserving deterministic predictions in unambiguous regions.

\begin{table*}[htp]
\caption{Comparison of the proposed Harmonizer Network against individually trained U-Net models (single-expert annotations, top), and personalized segmentation frameworks (bottom) on LIDC\textendash IDRI dataset.}
\centering
\resizebox{1.0\linewidth}{!}{
\begin{tabular}{c|cc|cc|ccccc}
\toprule[1.5pt]
\multirow{2}{*}{Method} & \multicolumn{2}{c|}{Diversity Performance} & \multicolumn{2}{c|}{Personalized Bounds (\%)}   & \multicolumn{5}{c}{Personalized Segmentation Performance (\%)} \\

\cline{2-10} & $GED$ $\downarrow$ & $Dice_{soft}\uparrow$ (\%) & $Dice_{max}\uparrow$ &   $Dice_{match}\uparrow$  &  $Dice_{A1}\uparrow$&  $Dice_{A2}\uparrow$ &  $Dice_{A3}\uparrow$ &  $Dice_{A4}\uparrow$ &  $Dice_{mean}\uparrow$        \\ \midrule
U-Net ($A_1$) &0.3062 &86.59 &\multicolumn{2}{c|}{\multirow{4}{*}{N/A}} &{87.80} &87.47 &85.49 &80.67 &85.36 \\
U-Net ($A_2$) &0.2459 &88.43 & & &87.16 &{89.08} &88.59 &85.15 &87.50\\
U-Net ($A_3$) &0.2436 &88.20 & & &85.29 &88.48 &{89.40} &87.20 &87.59\\
U-Net ($A_4$) &0.2962 &85.83 & & &80.80 &85.48 &88.22 &\textbf{88.90} &85.85\\ \midrule

Geometric-Structural \cite{wang2025noisy}& -& -& -& - &88.53&  87.98&  88.33&  88.16&  88.25\\
Prob. U-Net \cite{kohl2018probabilistic} &0.2168 &88.80 &88.87 &88.81 & -& -& -& -& \\
CM-Global \cite{tanno2019learning} &0.2432 &88.53 &87.51 &87.51 &86.13 &88.76 &88.99 &86.18 &87.51 \\
CM-Pixel \cite{zhang2020disentangling} &0.2407 &88.64 &87.72 &87.72 &85.99 &88.81 &89.31 &86.77 &87.72 \\
TAB \cite{liao2023transformer} &0.2322 &86.35 &87.11 &86.08 &85.00 &86.35 &86.77 &85.77 &85.97 \\
Pionono \cite{schmidt2023probabilistic}  &0.1502 &90.00 &90.10 &88.97 &87.94 &89.11 &89.55 &{88.76} &88.84 \\
D-Persona \cite{wu2024diversified} &{0.1444} &{90.31} &{90.38}&{89.17}&{88.54}&{89.50}&{90.03} &88.60 &{89.17}  \\
\rowcolor[HTML]{C8FFFD} 
Harmonizer Network &\textbf{0.1419} &\textbf{91.35} &\textbf{92.65}&\textbf{90.00}&\textbf{89.87}&\textbf{90.07}&\textbf{91.06} &{88.56} &\textbf{90.78}  \\
\bottomrule[1.5pt]
\end{tabular}}
\label{tab_lidc}
\end{table*}

\begin{table*}[htp]
\caption{Comparison of the proposed Harmonizer Network against individually trained U-Net models (single-expert annotations, top), and personalized segmentation frameworks (bottom) on NPC-170 dataset.}
\centering
\resizebox{1.0\linewidth}{!}{
\begin{tabular}{c|cc|cc|ccccc}
\toprule[1.5pt]
\multirow{2}{*}{Method} & \multicolumn{2}{c|}{Diversity Performance} & \multicolumn{2}{c|}{Personalized Bounds (\%)}   & \multicolumn{5}{c}{Personalized Segmentation Performance (\%)} \\

\cline{2-10} & $GED$ $\downarrow$ & $Dice_{soft}\uparrow$ (\%) & $Dice_{max}\uparrow$ &   $Dice_{match}\uparrow$  &  $Dice_{A1}\uparrow$&  $Dice_{A2}\uparrow$ &  $Dice_{A3}\uparrow$ &  $Dice_{A4}\uparrow$ &  $Dice_{mean}\uparrow$        \\ \midrule
U-Net ($A_1$) &0.4322 &78.65 &\multicolumn{2}{c|}{\multirow{4}{*}{N/A}} &{87.02} &73.68 &73.15 &77.05 &77.85 \\
U-Net ($A_2$) &0.4520 &76.45 & & &77.70 &{79.25} &74.10 &74.05 &76.35 \\
U-Net ($A_3$) &0.4515 &76.75 & & &76.35 &76.05 &{79.15} &74.85 &76.70 \\
U-Net ($A_4$) &0.4600 &77.80 & & &80.45 &76.10 &77.20 &{79.85} &78.90\\ \midrule
Geometric-Structural \cite{wang2025noisy}& -& -& -& - &82.37 &78.62 &77.64 &78.35 &79.24 \\
Prob. U-Net \cite{kohl2018probabilistic}  &0.3528 &81.19 &84.19 &80.13 & -& -& -& -& - \\
CM-Global \cite{tanno2019learning} &0.3609 &81.31 &\textbf{85.39} &80.13 &84.78 &77.60 &77.99 &{80.15} &80.13 \\
CM-Pixel \cite{zhang2020disentangling} &0.4523 &78.09 &81.20 &76.77 &80.47 &74.79 &74.06 &77.77 &76.77 \\
TAB \cite{liao2023transformer} &{0.2900} &81.25 &81.60 &80.02 &83.10 &78.10 &79.50 &76.50 &79.30 \\
Pionono \cite{schmidt2023probabilistic}  &0.33348 &81.66 &85.00 &80.32 &85.10 &77.59 &78.27 &80.31 &80.32 \\
D-Persona\cite{wu2024diversified} &0.2970 &82.30 &81.60 &80.50 &84.20 &\textbf{79.70} &\textbf{80.50} &77.20 &80.40 \\
\rowcolor[HTML]{C8FFFD} 
Harmonizer Network &\textbf{0.2685} &\textbf{83.10} &{84.46} &\textbf{81.63} &\textbf{85.79} &{79.30} &{80.40} &\textbf{81.03} &\textbf{81.63} \\
\bottomrule[1.5pt]
\end{tabular}}
\label{tab_npc}
\end{table*}
\paragraph{Personalization segmentation:}
When augmented with rater-aware frequency prompts in phase II, the Harmonizer Network delivers further improvements over all personalized baselines (Tables \ref{tab_lidc} and \ref{tab_npc}). On LIDC-IDRI, our model achieves the best personalized bounds (Dice${\text{max}}$ = 92.65\%, Dice${\text{match}}$ = 90.0\%) and balanced per-rater performance, outperforming D-Persona by roughly +1.61 pp in Dice${\text{mean}}$. On NPC-170, a more challenging multi-modal dataset, our method surpasses both transformer-based TAB \cite{liao2023transformer} and probabilistic Pionono \cite{schmidt2023probabilistic}, reaching 81.63\% mean Dice despite pronounced inter-rater divergence. Unlike D-Persona, which conditions expert prompts on spatial features susceptible to residual acquisition noise, our frequency-prompt module operates on harmonized spectral representations, allowing it to isolate stylistic boundary cues (e.g., sharpness, textural sensitivity) without disrupting structural consistency. This separation yields personalized masks that are both anatomically aligned and rater-specific: the gap between Dice${\text{max}}$ and Dice$_{\text{match}}$ remains narrow, indicating that each individualized prediction is genuinely tailored rather than a random sample fitting by chance. Qualitative visualization (\ref{fig:resmain}) further shows that our personalized outputs adaptively refine boundary thickness and contrast according to each annotator’s style, validating that the proposed frequency-aware conditioning achieves a more faithful and interpretable form of expert personalization.

Moreover, the top blocks in Tables~\ref{tab_lidc} and~\ref{tab_npc} report four independently trained U-Nets, each tuned to a single annotator. While each model peaks on its own rater (bold entries), it underperforms on others and lacks distributional coverage. Our method, by contrast, produces (i) a calibrated set of hypotheses that aligns with the empirical annotation distribution and (ii) personalized predictions that respect individual styles without training one network per reader.

\section*{Ablation Study and Acknowledgments}
Ablation is provided in the supplementary material. This work was funded by the German Research Foundation
(DFG) – project number 455548460.

\section{Conclusion}
We introduced the Harmonizer Network, a novel probabilistic multi-rater medical image segmentation framework. To address literature gap, the first stage employs a Noise Harmonizer with GED-based alignment to model artifact-invariant latent distributions, while the second stage integrates frequency-domain prompts to capture annotator-specific variations. Evaluations on LIDC-IDRI and NPC-170 show consistent improvements over probabilistic and personalized methods in distributional fit and personalized accuracy. The model concentrates uncertainty in ambiguous regions, maintains balanced per-rater adaptation, and remains robust to noise and blur (see supplementary), improving segmentation reliability and interpretability.

{
    \small
    \bibliographystyle{ieeenat_fullname}
    \bibliography{main}

@String(CVPR= {IEEE Conf. Comput. Vis. Pattern Recog.})

@String(ICCV= {Int. Conf. Comput. Vis.})

@String(ICLR = {Int. Conf. Learn. Represent.})

@String(AAAI = {AAAI})

@String(CVPR  = {CVPR})

@String(ICCV  = {ICCV})

@String(ICLR  = {ICLR})

@inproceedings{lourencco2021using,
  title={Using soft labels to model uncertainty in medical image segmentation},
  author={Louren{\c{c}}o-Silva, Jo{\~a}o and Oliveira, Arlindo L},
  booktitle={International MICCAI brainlesion workshop},
  pages={585--596},
  year={2021},
  organization={Springer}
}

@inproceedings{mirikharaji2021d,
  title={D-lema: Deep learning ensembles from multiple annotations-application to skin lesion segmentation},
  author={Mirikharaji, Zahra and Abhishek, Kumar and Izadi, Saeed and Hamarneh, Ghassan},
  booktitle={Proceedings of the IEEE/CVF Conference on Computer Vision and Pattern Recognition},
  pages={1837--1846},
  year={2021}
}

@article{wu2024calibrate,
  title={Calibrate the inter-observer segmentation uncertainty via diagnosis-first principle},
  author={Wu, Junde and Zhang, Yu and Fang, Huihui and Duan, Lixin and Tan, Mingkui and Yang, Weihua and Wang, Chunhui and Liu, Huiying and Jin, Yueming and Xu, Yanwu},
  journal={IEEE Transactions on Medical Imaging},
  volume={43},
  number={9},
  pages={3331--3342},
  year={2024},
  publisher={IEEE}
}

@article{Zhang2025DiffOSeg,
  author  = {Han Zhang and Xiangde Luo and Yong Chen and Kang Li},
  title   = {DiffOSeg: Omni Medical Image Segmentation via Multi-Expert Collaboration Diffusion Model},
  journal = {arXiv preprint arXiv:2507.13087},
  year    = {2025}
}

@inproceedings{baumgartner2019phiseg,
  title={Phiseg: Capturing uncertainty in medical image segmentation},
  author={Baumgartner, Christian F and Tezcan, Kerem C and Chaitanya, Krishna and H{\"o}tker, Andreas M and Muehlematter, Urs J and Schawkat, Khoschy and Becker, Anton S and Donati, Olivio and Konukoglu, Ender},
  booktitle={International Conference on Medical Image Computing and Computer-Assisted Intervention},
  pages={119--127},
  year={2019},
  organization={Springer}
}

@inproceedings{rahman2023ambiguous,
  title={Ambiguous medical image segmentation using diffusion models},
  author={Rahman, Aimon and Valanarasu, Jeya Maria Jose and Hacihaliloglu, Ilker and Patel, Vishal M},
  booktitle={Proceedings of the IEEE/CVF conference on computer vision and pattern recognition},
  pages={11536--11546},
  year={2023}
}

@article{warfield2004simultaneous,
  title={Simultaneous truth and performance level estimation (STAPLE): an algorithm for the validation of image segmentation},
  author={Warfield, Simon K and Zou, Kelly H and Wells, William M},
  journal={IEEE transactions on medical imaging},
  volume={23},
  number={7},
  pages={903--921},
  year={2004},
  publisher={IEEE}
}

@inproceedings{wu2024diversified,
  title={Diversified and personalized multi-rater medical image segmentation},
  author={Wu, Yicheng and Luo, Xiangde and Xu, Zhe and Guo, Xiaoqing and Ju, Lie and Ge, Zongyuan and Liao, Wenjun and Cai, Jianfei},
  booktitle={Proceedings of the IEEE/CVF Conference on Computer Vision and Pattern Recognition},
  pages={11470--11479},
  year={2024}
}

@inproceedings{guan2018said,
  title={Who said what: Modeling individual labelers improves classification},
  author={Guan, Melody and Gulshan, Varun and Dai, Andrew and Hinton, Geoffrey},
  booktitle={Proceedings of the AAAI conference on artificial intelligence},
  volume={32},
  number={1},
  year={2018}
}

@inproceedings{azad2023laplacian,
  title={Laplacian-former: Overcoming the limitations of vision transformers in local texture detection},
  author={Azad, Reza and Kazerouni, Amirhossein and Azad, Babak and Khodapanah Aghdam, Ehsan and Velichko, Yury and Bagci, Ulas and Merhof, Dorit},
  booktitle={International Conference on Medical Image Computing and Computer-Assisted Intervention},
  pages={736--746},
  year={2023},
  organization={Springer}
}

@article{guo2023visual,
  title={Visual attention network},
  author={Guo, Meng-Hao and Lu, Cheng-Ze and Liu, Zheng-Ning and Cheng, Ming-Ming and Hu, Shi-Min},
  journal={Computational visual media},
  volume={9},
  number={4},
  pages={733--752},
  year={2023},
  publisher={TUP}
}

@inproceedings{Wang2024MultiRaterPrompt,
  author    = {Jinhong Wang and Yi Cheng and Jintai Chen and Hongxia Xu and Danny Chen and Jian Wu},
  title     = {Multi-rater Prompting for Ambiguous Medical Image Segmentation},
  booktitle = {IEEE International Conference on Bioinformatics and Biomedicine (BIBM) 2024},
  year      = {2024}
}

@article{cheng2023tax,
  title={TAX: Tendency-and-assignment explainer for semantic segmentation with multi-annotators},
  author={Cheng, Yuan-Chia and Shiau, Zu-Yun and Yang, Fu-En and Wang, Yu-Chiang Frank},
  journal={arXiv preprint arXiv:2302.09561},
  year={2023}
}

@inproceedings{Zepf2023That,
  author    = {Kilian Zepf and Eike Petersen and Jes Frellsen and Aasa Feragen},
  title     = {That Label's Got Style: Handling Label Style Bias for Uncertain Image Segmentation},
  booktitle = {Proc. International Conference on Learning Representations (ICLR)},
  year      = {2023}
}

@article{wang2025noisy,
  title={From Noisy Labels to Intrinsic Structure: A Geometric-Structural Dual-Guided Framework for Noise-Robust Medical Image Segmentation},
  author={Wang, Tao and Zhang, Zhenxuan and Zhou, Yuanbo and Zhang, Xinlin and Chen, Yuanbin and Tan, Tao and Yang, Guang and Tong, Tong},
  journal={arXiv preprint arXiv:2509.02419},
  year={2025}
}

@article{wang2025learning,
  title={Learning robust medical image segmentation from multi-source annotations},
  author={Wang, Yifeng and Luo, Luyang and Wu, Mingxiang and Wang, Qiong and Chen, Hao},
  journal={Medical Image Analysis},
  volume={101},
  pages={103489},
  year={2025},
  publisher={Elsevier}
}

@article{hu2023inter,
  title={Inter-rater uncertainty quantification in medical image segmentation via rater-specific bayesian neural networks},
  author={Hu, Qingqiao and Wang, Hao and Luo, Jing and Luo, Yunhao and Zhangg, Zhiheng and Kirschke, Jan S and Wiestler, Benedikt and Menze, Bjoern and Zhang, Jianguo and Li, Hongwei Bran},
  journal={arXiv preprint arXiv:2306.16556},
  year={2023}
}

@article{zhang2020disentangling,
  title={Disentangling human error from ground truth in segmentation of medical images},
  author={Zhang, Le and Tanno, Ryutaro and Xu, Mou-Cheng and Jin, Chen and Jacob, Joseph and Cicarrelli, Olga and Barkhof, Frederik and Alexander, Daniel},
  journal={Advances in Neural Information Processing Systems},
  volume={33},
  pages={15750--15762},
  year={2020}
}

@inproceedings{ye2023confidence,
  title={Confidence contours: Uncertainty-aware annotation for medical semantic segmentation},
  author={Ye, Andre and Chen, Quan Ze and Zhang, Amy},
  booktitle={Proceedings of the AAAI Conference on Human Computation and Crowdsourcing},
  volume={11},
  pages={186--197},
  year={2023}
}

@article{li2024qubiq,
  title={Qubiq: Uncertainty quantification for biomedical image segmentation challenge},
  author={Li, Hongwei Bran and Navarro, Fernando and Ezhov, Ivan and Bayat, Amirhossein and Das, Dhritiman and Kofler, Florian and Shit, Suprosanna and Waldmannstetter, Diana and Paetzold, Johannes C and Hu, Xiaobin and others},
  journal={arXiv preprint arXiv:2405.18435},
  year={2024}
}

@article{kohl2018probabilistic,
  title={A probabilistic u-net for segmentation of ambiguous images},
  author={Kohl, Simon and Romera-Paredes, Bernardino and Meyer, Clemens and De Fauw, Jeffrey and Ledsam, Joseph R and Maier-Hein, Klaus and Eslami, SM and Jimenez Rezende, Danilo and Ronneberger, Olaf},
  journal={Advances in neural information processing systems},
  volume={31},
  year={2018}
}

@article{monteiro2020stochastic,
  title={Stochastic segmentation networks: Modelling spatially correlated aleatoric uncertainty},
  author={Monteiro, Miguel and Le Folgoc, Lo{\"\i}c and Coelho de Castro, Daniel and Pawlowski, Nick and Marques, Bernardo and Kamnitsas, Konstantinos and Van der Wilk, Mark and Glocker, Ben},
  journal={Advances in neural information processing systems},
  volume={33},
  pages={12756--12767},
  year={2020}
}

@article{cheplygina2019notsosupervised,
  title={Not-so-supervised: a survey of semi-supervised, multi-instance, and transfer learning in medical image analysis},
  author={Cheplygina, Veronika and de Bruijne, Marleen and Pluim, Josien PW},
  journal={Medical Image Analysis},
  volume={54},
  pages={280--296},
  year={2019},
  publisher={Elsevier},
  doi={10.1016/j.media.2019.03.009}
}

@inproceedings{kumari2025annotation,
  title={Annotation Ambiguity Aware Semi-Supervised Medical Image Segmentation},
  author={Kumari, Suruchi and Singh, Pravendra},
  booktitle={Proceedings of the Computer Vision and Pattern Recognition Conference},
  pages={10404--10413},
  year={2025}
}

@inproceedings{ji2021learning,
  title={Learning calibrated medical image segmentation via multi-rater agreement modeling},
  author={Ji, Wei and Yu, Shuang and Wu, Junde and Ma, Kai and Bian, Cheng and Bi, Qi and Li, Jingjing and Liu, Hanruo and Cheng, Li and Zheng, Yefeng},
  booktitle=CVPR,
  pages={12341--12351},
  year={2021}
}

@article{wang2023medical,
  title={Medical matting: Medical image segmentation with uncertainty from the matting perspective},
  author={Wang, Lin and Ye, Xiufen and Ju, Lie and He, Wanji and Zhang, Donghao and Wang, Xin and Huang, Yelin and Feng, Wei and Song, Kaimin and Ge, Zongyuan},
  journal={Computers in Biology and Medicine},
  volume={158},
  pages={106714},
  year={2023}
}

@inproceedings{schmidt2023probabilistic,
  title={Probabilistic Modeling of Inter-and Intra-observer Variability in Medical Image Segmentation},
  author={Schmidt, Arne and Morales-{\'A}lvarez, Pablo and Molina, Rafael},
  booktitle=ICCV,
  pages={21097--21106},
  year={2023}
}

@inproceedings{liao2023transformer,
  title={Transformer-based annotation bias-aware medical image segmentation},
  author={Liao, Zehui and Hu, Shishuai and Xie, Yutong and Xia, Yong},
  booktitle={International conference on medical image computing and computer-assisted intervention},
  pages={24--34},
  year={2023}
}

@article{kohl2019hierarchical,
  title={A hierarchical probabilistic u-net for modeling multi-scale ambiguities},
  author={Kohl, Simon AA and Romera-Paredes, Bernardino and Maier-Hein, Klaus H and Rezende, Danilo Jimenez and Eslami, SM and Kohli, Pushmeet and Zisserman, Andrew and Ronneberger, Olaf},
  journal={arXiv preprint arXiv:1905.13077},
  year={2019}
}

@article{bellemare2017cramer,
  title={The cramer distance as a solution to biased wasserstein gradients},
  author={Bellemare, Marc G and Danihelka, Ivo and Dabney, Will and Mohamed, Shakir and Lakshminarayanan, Balaji and Hoyer, Stephan and Munos, R{\'e}mi},
  journal={arXiv preprint arXiv:1705.10743},
  year={2017}
}

@inproceedings{wang2019symmetric,
  title={Symmetric cross entropy for robust learning with noisy labels},
  author={Wang, Yisen and Ma, Xingjun and Chen, Zaiyi and Luo, Yuan and Yi, Jinfeng and Bailey, James},
  booktitle=ICCV,
  pages={322--330},
  year={2019}
}

@inproceedings{tanno2019learning,
  title={Learning from noisy labels by regularized estimation of annotator confusion},
  author={Tanno, Ryutaro and Saeedi, Ardavan and Sankaranarayanan, Swami and Alexander, Daniel C and Silberman, Nathan},
  booktitle=CVPR,
  pages={11244--11253},
  year={2019}
}

@article{valiuddin2024investigating,
  title={Investigating and Improving Latent Density Segmentation Models for Aleatoric Uncertainty Quantification in Medical Imaging},
  author={Valiuddin, MM Amaan and Viviers, Christiaan GA and Van Sloun, Ruud JG and De With, Peter HN and van der Sommen, Fons},
  journal={IEEE Transactions on Medical Imaging},
  year={2024},
  publisher={IEEE}
}

@article{armato2011lidc,
  title={The lung image database consortium (LIDC) and image database resource initiative (IDRI): a completed reference database of lung nodules on CT scans},
  author={Armato, Samuel G and McLennan, Geoffrey and Bidaut, Luc and McNitt-Gray, Michael F and Meyer, Charles R and Reeves, Anthony P and Zhao, Binsheng and Aberle, Denise R and Henschke, Claudia I and Hoffman, Eric A and others},
  journal={Medical physics},
  volume={38},
  number={2},
  pages={915--931},
  year={2011},
  publisher={Wiley Online Library}
}

@inproceedings{czolbe2021useful,
  title={Is Segmentation Uncertainty Useful?},
  author={Czolbe, Steffen and Arnavaz, Kasra and Krause, Oswin and Feragen, Aasa},
  booktitle={Information Processing in Medical Imaging (IPMI)},
  pages={715--726},
  volume={12729},
  year={2021},
  organization={Springer}
}

@inproceedings{jha2019kvasir,
  title={Kvasir-seg: A segmented polyp dataset},
  author={Jha, Debesh and Smedsrud, Pia H and Riegler, Michael A and Halvorsen, P{\aa}l and De Lange, Thomas and Johansen, Dag and Johansen, H{\aa}vard D},
  booktitle={International conference on multimedia modeling},
  pages={451--462},
  year={2019},
  organization={Springer}
}

@article{zhao2024ultrasound,
  title={Ultrasound nodule segmentation using asymmetric learning with simple clinical annotation},
  author={Zhao, Xingyue and Li, Zhongyu and Luo, Xiangde and Li, Peiqi and Huang, Peng and Zhu, Jianwei and Liu, Yang and Zhu, Jihua and Yang, Meng and Chang, Shi and others},
  journal={IEEE Transactions on Circuits and Systems for Video Technology},
  volume={34},
  number={10},
  pages={9010--9023},
  year={2024},
  publisher={IEEE}
}

@inproceedings{liu2022adaptive,
  title={Adaptive early-learning correction for segmentation from noisy annotations},
  author={Liu, Sheng and Liu, Kangning and Zhu, Weicheng and Shen, Yiqiu and Fernandez-Granda, Carlos},
  booktitle={Proceedings of the IEEE/CVF conference on computer vision and pattern recognition},
  pages={2606--2616},
  year={2022}
}

@inproceedings{xia2020robust,
  title={Robust early-learning: Hindering the memorization of noisy labels},
  author={Xia, Xiaobo and Liu, Tongliang and Han, Bo and Gong, Chen and Wang, Nannan and Ge, Zongyuan and Chang, Yi},
  booktitle={International conference on learning representations},
  year={2020}
}

@article{han2018co,
  title={Co-teaching: Robust training of deep neural networks with extremely noisy labels},
  author={Han, Bo and Yao, Quanming and Yu, Xingrui and Niu, Gang and Xu, Miao and Hu, Weihua and Tsang, Ivor and Sugiyama, Masashi},
  journal={Advances in neural information processing systems},
  volume={31},
  year={2018}
}

@inproceedings{wei2020combating,
  title={Combating noisy labels by agreement: A joint training method with co-regularization},
  author={Wei, Hongxin and Feng, Lei and Chen, Xiangyu and An, Bo},
  booktitle={Proceedings of the IEEE/CVF conference on computer vision and pattern recognition},
  pages={13726--13735},
  year={2020}
}

@article{fang2023reliable,
  title={Reliable mutual distillation for medical image segmentation under imperfect annotations},
  author={Fang, Chaowei and Wang, Qian and Cheng, Lechao and Gao, Zhifan and Pan, Chengwei and Cao, Zhen and Zheng, Zhaohui and Zhang, Dingwen},
  journal={IEEE Transactions on Medical Imaging},
  volume={42},
  number={6},
  pages={1720--1734},
  year={2023},
  publisher={IEEE}
}

@inproceedings{li2021superpixel,
  title={Superpixel-guided iterative learning from noisy labels for medical image segmentation},
  author={Li, Shuailin and Gao, Zhitong and He, Xuming},
  booktitle={International Conference on Medical Image Computing and Computer-Assisted Intervention},
  pages={525--535},
  year={2021},
  organization={Springer}
}

@InProceedings{guo17a,
  title     = {On Calibration of Modern Neural Networks},
  author    = {Chuan Guo and Geoff Pleiss and Yu Sun and Kilian Q. Weinberger},
  booktitle = {Proceedings of the 34th International Conference on Machine Learning (ICML)},
  series    = {Proceedings of Machine Learning Research},
  volume    = {70},
  pages     = {1321--1330},
  year      = {2017},
  url       = {https://proceedings.mlr.press/v70/guo17a/guo17a.pdf}
}

@Article{steyerberg10,
  title   = {Assessing the performance of prediction models: a framework for some traditional and novel measures},
  author  = {Ewout W. Steyerberg and Andrew J. Vickers and Nancy R. Cook and Thomas Gerds and Mithat Gonen and Nancy Obuchowski and Michael J. Pencina and Michael W. Kattan},
  journal = {Epidemiology},
  volume  = {21},
  number  = {1},
  pages   = {128--138},
  year    = {2010},
  url     = {https://www.ncbi.nlm.nih.gov/pmc/articles/PMC3575184/}
}
}

\newpage
\appendix
\section*{Supplementary Material}
To further support our experimental results, we provide additional empirical evidence and extended analyses validating the proposed Harmonizer Network. 
This supplementary material includes detailed ablations on hyperparameters, GED behavior, relationships between uncertainty and correctness, alignment of uncertainty with rater disagreement, per-rater calibration, lesion-size robustness, and the effects of noise. 
These studies demonstrate that the observed performance gains arise from principled model design rather than dataset-specific tuning. 
Additional visualizations on the LIDC-IDRI, NPC-170, and Kvasir datasets illustrate consistent boundary quality, realistic latent-space diversity, and faithful personalization of individual rater styles. 
Robustness experiments under severe noise confirm the stability of the harmonized latent space, while frequency-domain visualizations reveal how high-frequency prompts enhance boundary-specific details. 
Finally, cross-dataset evaluations and efficiency analyses verify that the method generalizes well, remains computationally lightweight, and delivers superior performance even under challenging noisy-label conditions.

\section{Selection of Hyper-parameters}
\label{sec:hyperparams}
To ensure a fair and reproducible evaluation, all hyperparameters governing the composite loss in Eq.~(8) of the main paper were systematically tuned through one-fold validation on the LIDC--IDRI dataset and subsequently fixed across all experiments and datasets. Specifically, we optimized the balance among the segmentation loss ($\mathcal{L}_{\text{seg}}$), KL divergence term, harmonization penalty, and GED-based regularizer. The weighting coefficients $\lambda_{\text{KL}}$, $\lambda_{\text{harm}}$, and $\lambda_{\text{GED}}$ were empirically selected to achieve stable convergence and the best trade-off between reconstruction fidelity, latent regularity, and distributional alignment. 

During tuning, we performed a grid search around $\lambda_{\text{KL}} \in [1,5]\!\times\!10^{-3}$, $\lambda_{\text{harm}} \in [1,5]\!\times\!10^{-4}$, and $\lambda_{\text{GED}} \in [0.5,2.0]$, evaluating each configuration by the validation GED and soft-Dice scores. The optimal configuration ($\lambda_{\text{KL}} = 2\!\times\!10^{-3}$, $\lambda_{\text{harm}} = 3\!\times\!10^{-4}$, $\lambda_{\text{GED}} = 1.0$) was adopted for all subsequent experiments, including NPC--170, without further modification. This strategy provided consistent convergence behaviour, stable uncertainty calibration, and ensured that improvements observed in cross-dataset evaluations stemmed from the model design rather than dataset-specific hyperparameter tuning.

\section{GED vs.\ Sample Count}
\label{sec:ged_vs_samples}
\begin{figure}[t]
    \centering
    \includegraphics[width=1\linewidth]{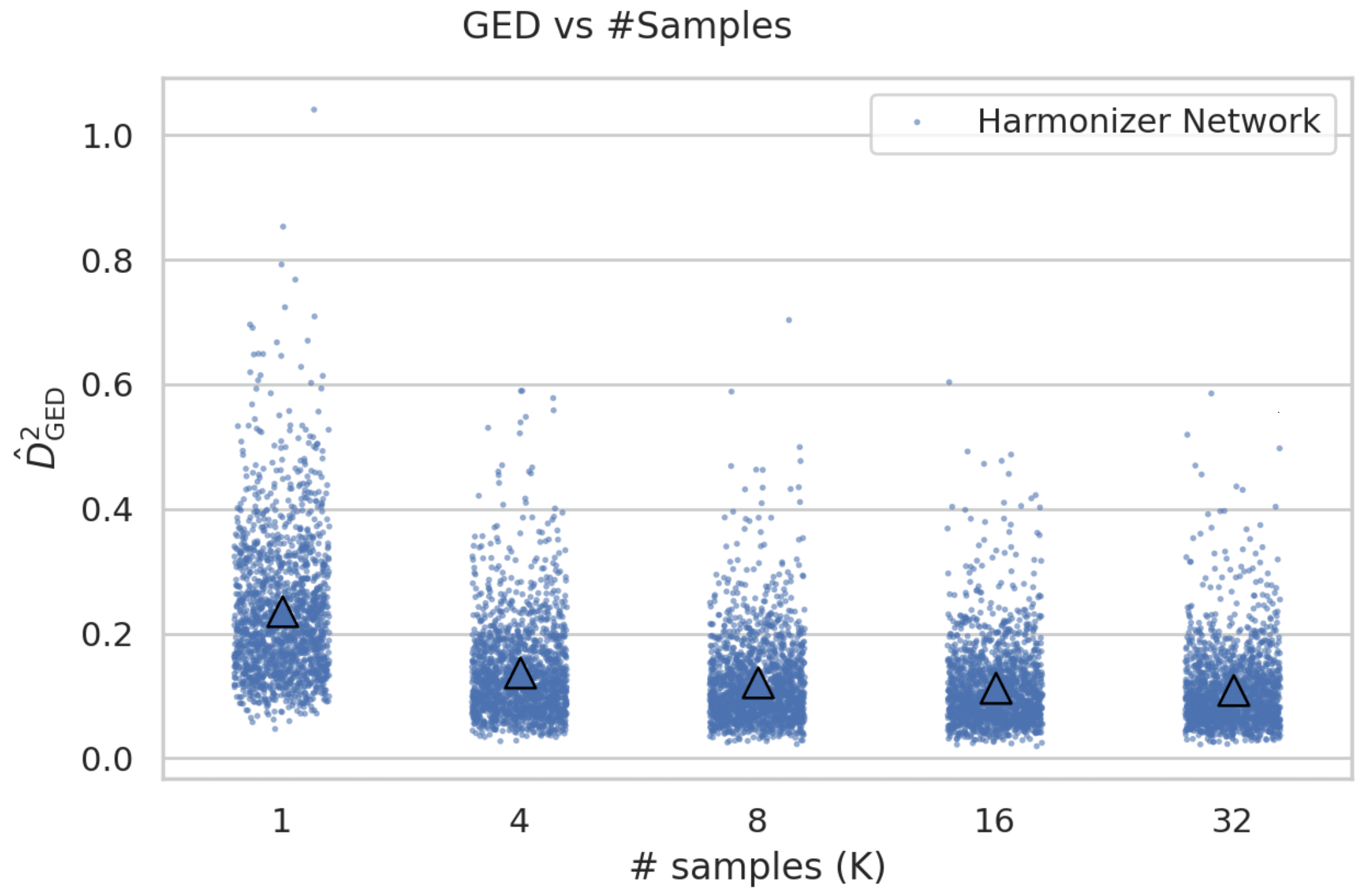}
    \caption{
    Impact of the proposed GED loss on LIDC. 
    GED is reported as a function of the number of generated samples \(K\) (lower is better). 
    Models trained with the GED loss demonstrate consistently lower discrepancy and reach saturation with fewer samples, indicating more efficient coverage of the underlying rater distribution.
    }
    \label{fig:ged_vs_samples}
\end{figure}

To examine how well the proposed model approximates the empirical annotation distribution as the number of generated samples increases, we analyze the GED as a function of the number of prior samples $K$, following the standard evaluation protocol introduced in Probabilistic U\textendash Net~\cite{kohl2018probabilistic}. For each test image and $K \in \{1, 4, 8, 16, 32\}$, we draw $K$ stochastic segmentations predictions to compute GED between the predicted and annotated sets using the soft-Dice distance from GED loss, and report the mean~$\pm$~standard deviation across folds. As expected, GED decreases monotonically with increasing $K$, reflecting improved coverage of annotation variability, and stabilizes once additional samples contribute little new diversity. Incorporating the GED regularizer during training leads to a consistent downward shift in the GED curve and earlier saturation, showing that the learned prior allocates its variability primarily to ambiguous regions while remaining deterministic elsewhere. Formally, optimizing $\mathcal{L}_{\text{GED}} = 2\,\mathbb{E}[d(P,A)] - \mathbb{E}[d(P,P')]$, with $d(\cdot,\cdot)$ denoting the soft-Dice distance, simultaneously reduces cross-set discrepancies and maintains controlled intra-set diversity, yielding a more calibrated posterior. In practice, GED computation scales linearly with $K$ for $d(P,A)$ and quadratically for $d(P,P')$; we therefore cap $K$ at~32 and subsample pairs for $d(P,P')$ without affecting observed trends. As shown in Figure~\ref{fig:ged_vs_samples}, the Harmonizer prior exhibits consistently lower GED values and faster convergence than D\textendash Persona~\cite{wu2024diversified}, confirming its superior distributional alignment and sample efficiency in capturing clinically meaningful annotation diversity.

\section{Uncertainty vs.\ Prediction Correctness}
\label{sec:uncert_vs_correctness}

We assess whether the model's pixelwise predictive uncertainty is indicative of prediction correctness, following the evaluation protocol from \cite{czolbe2021useful}. For each test image, we generate $K=16$ stochastic segmentations from the learned prior and compute the mean per-pixel foreground probability $\bar{p}(x)$ across samples. Uncertainty is quantified as normalized binary entropy ($H(\hat{p})$)of the predictive mean:
\[
H(\hat{p})(x) = -\,\bar{p}(x)\log_2 \bar{p}(x) - (1-\bar{p}(x))\log_2(1-\bar{p}(x)),
\label{eq:entropy}
\]
which yields values in the range $[0, 1]$.

To establish ground truth in the presence of multiple annotators, we average per-pixel rater masks to form a consensus map, and retain only those pixels with unanimous agreement (i.e., consensus label exactly 0 or 1), filtering out ambiguous regions unless noted otherwise. Thresholding $\bar{p}(x)$ at 0.5, we classify each retained pixel into one of four correctness categories: True Positive (TP), False Positive (FP), False Negative (FN), or True Negative (TN).

\begin{figure}[t]
  \centering
  \includegraphics[width=0.94\linewidth]{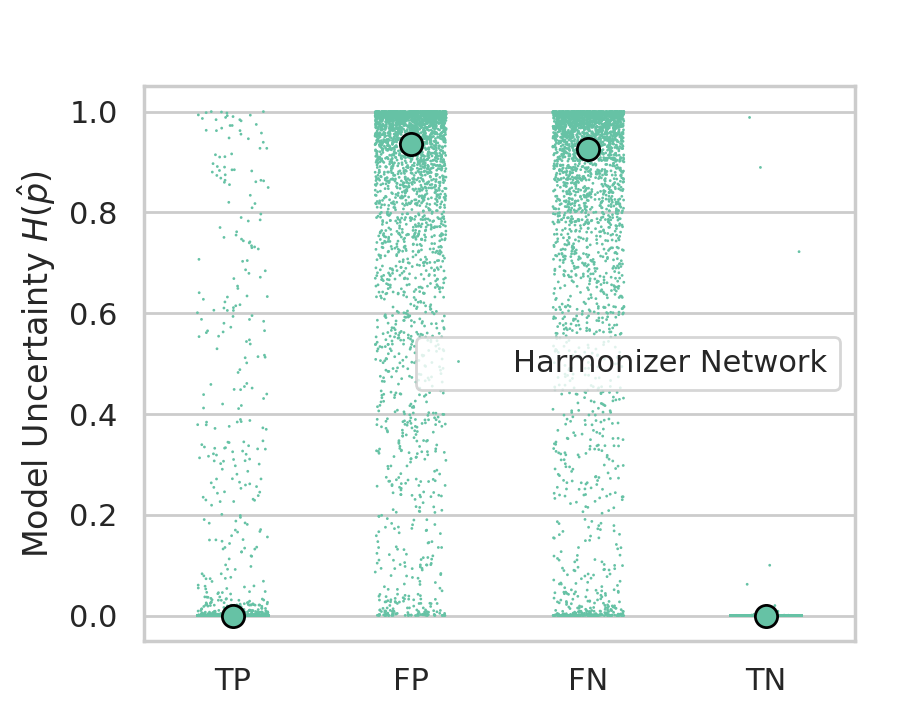}
  \vspace{-0.5em}
  \caption{Pixelwise uncertainty by correctness (TP/FP/FN/TN). Dots are individual pixels; medians are circled.}
  \label{fig:uncert_vs_correctness}
\end{figure}

Figure~\ref{fig:uncert_vs_correctness} presents a scatter plot of entropy values grouped by TP/FP/FN/TN classification. Each dot represents an individual pixel, and per-group medians are highlighted with circled markers. As expected for a well-calibrated model, TN and TP pixels exhibit near-zero uncertainty, while FP and FN pixels cluster around high entropy, with medians approaching 1.0. This sharp separation between correct and incorrect predictions confirms that the model is confident when accurate, and uncertain when wrong, an essential property for reliable downstream usage.

Interestingly, a small subset of true positives shows elevated entropy. These typically lie near lesion boundaries or within low-contrast regions, where even correct predictions tend to be probabilistic, yielding $\bar{p}(x) \approx 0.5$ and thus high entropy. In contrast, TN pixels, corresponding to background, consistently show minimal uncertainty, which can be attributed to their spatial homogeneity, larger area coverage, and stronger rater agreement.

This pattern persists even in the presence of noisy boundaries: although some entropy values are elevated near object edges due to genuine inter-rater disagreement, the group-wise medians remain cleanly separated. This suggests that uncertainty is not uniformly inflated but is instead tightly coupled to the model's confidence in its predictions.

In practical terms, this behavior enables automatic pixel- or region-level triage. High-uncertainty areas align with prediction errors (FP/FN), making them candidates for clinician review or further processing, while low-uncertainty regions (TP/TN) indicate reliable predictions. The observed asymmetry, where some TPs have high uncertainty but TNs rarely do, is consistent with the nature of the task: foreground lesions are more variable and subject to subjective interpretation, while background regions are more stable.

Hence, the model's entropy-based uncertainty estimates are both diverse and diagnostic. They not only reflect inter-rater ambiguity, but also serve as a practical signal for identifying erroneous predictions, reinforcing their interpretability and utility in multi-rater medical segmentation workflows.

\section{Size-Stratified Robustness}
\label{subsec:size_robustness}

To examine whether our model maintains consistent performance across lesions of different sizes, we perform a size-stratified evaluation, motivated by the well-known challenge that small lesions are typically harder to segment due to their limited spatial extent, while large lesions may introduce structural variability. A model that overfits to a dominant lesion size in the training distribution could fail to generalize across the full spectrum of cases seen in practice. This ablation explicitly tests for such scale-related biases.

\begin{figure}[t]
  \centering
  \includegraphics[width=\columnwidth]{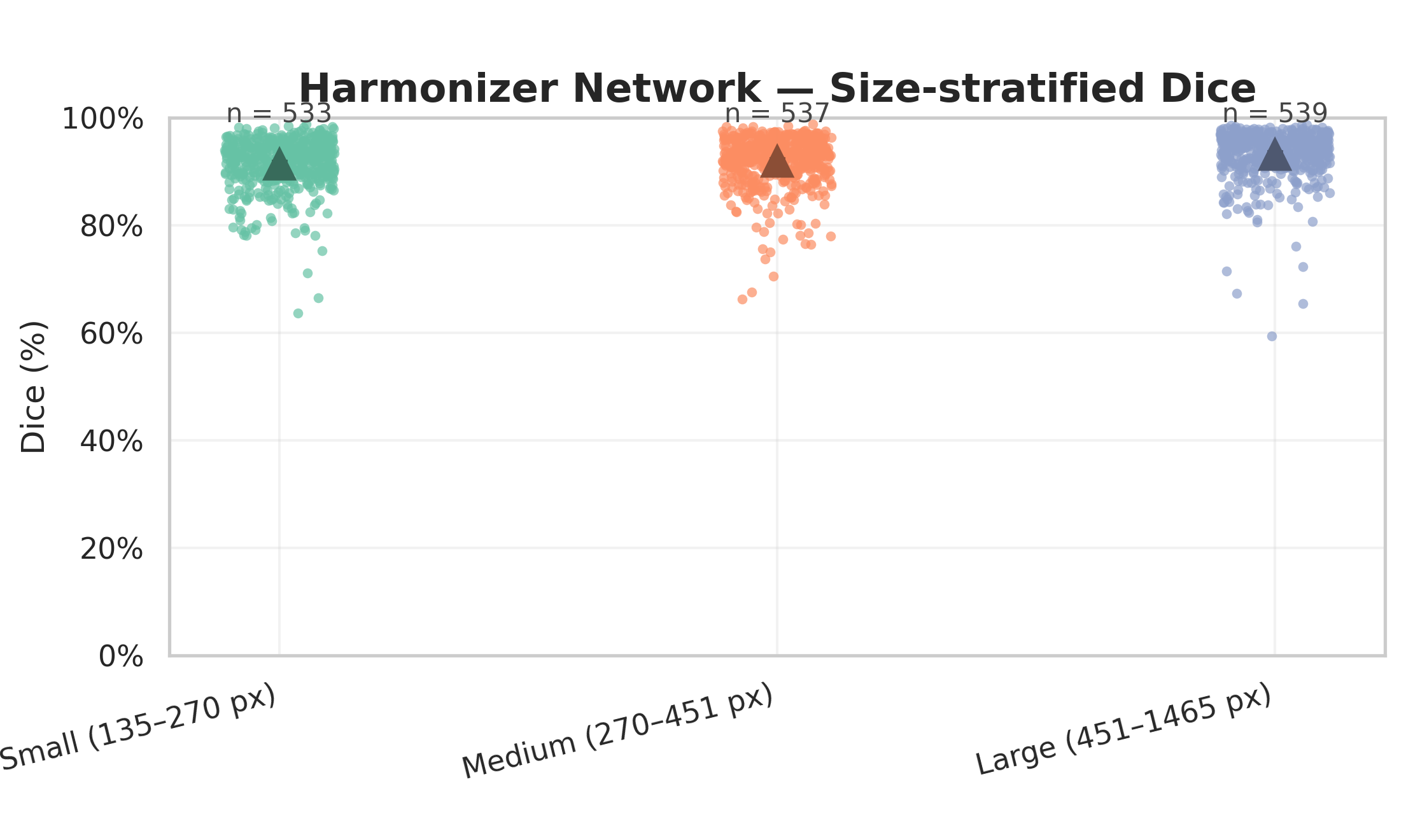}
  \caption{\textbf{Size-stratified Dice.} Each dot represents a single test case, and large triangles mark the mean Dice per bin. \emph{n} indicates the number of cases per bin. Similar means and variances across bins confirm that model performance is consistent across lesion sizes.}
  \label{fig:size_bins}
\end{figure}

For each test image, we compute a consensus mask by averaging expert annotations and thresholding at 0.5 to obtain a binary foreground region. Lesion size is then defined as the pixel count of the resulting foreground mask. Using size quantiles computed on the test set, we partition all lesions into three equally populated bins: \textit{Small} (135–270 px), \textit{Medium} (270–451 px), and \textit{Large} (451–1465 px), with ranges shown on the x-axis of Figure~\ref{fig:size_bins}.

Within each bin, we compute the Dice similarity between the model’s predicted mask and the consensus annotation. Figure~\ref{fig:size_bins} plots all per-case Dice scores as individual points (with horizontal jitter for visibility), while large colored triangles represent the mean Dice within each bin. The total number of test cases in each group is annotated above the plots.

The results show high and consistent Dice scores across all three bins, with overlapping distributions and tightly clustered means. Importantly, the model performs robustly on small lesions, which are often more ambiguous due to faint boundaries, low contrast, and lower inter-rater agreement. The fact that performance does not drop for these more challenging cases suggests that our method is not biased toward larger or more prominent regions, and is capable of reliably segmenting lesions regardless of their scale.

This ablation confirms that our model demonstrates strong scale-invariance in segmentation accuracy, maintaining stable performance from small to large lesions. This robustness is critical for clinical applicability, where lesion size varies widely between patients and imaging scenarios.

\section{Model Uncertainty vs.\ Rater Uncertainty}
\label{subsec:uncert_vs_raters}

A key question in uncertainty modeling is whether the model’s predictive uncertainty arises from genuine ambiguity in expert annotations, or whether it simply reflects noise or internal instability. To address this, we examine how model uncertainty correlates with rater uncertainty, as measured by inter-expert agreement on the test set. The goal is to verify whether high model entropy coincides with regions where annotators themselves disagree, suggesting that uncertainty is interpretable and human-aligned rather than random.

For each test image, we sample $K$ stochastic predictions from the learned prior and compute pixelwise model uncertainty using the entropy formulation defined in Eq.~\ref{eq:entropy}, where $\bar{p}(x)$ denotes the mean per-pixel foreground probability, with entropy values normalized to the range $[0,1]$.

We then compute per-pixel rater uncertainty from the mean annotation probability $p_A$, obtained by averaging binary masks from all experts. Based on $p_A$, each pixel is assigned to one of three rater agreement regimes: 
agree ($p_A \approx 0$ or $1$), 
somewhat agree (intermediate values away from 0.5), 
and disagree ($p_A \approx 0.5$), using small tolerance margins around the endpoints and midpoint.

\begin{figure}[t]
  \centering
  \includegraphics[width=1\linewidth]{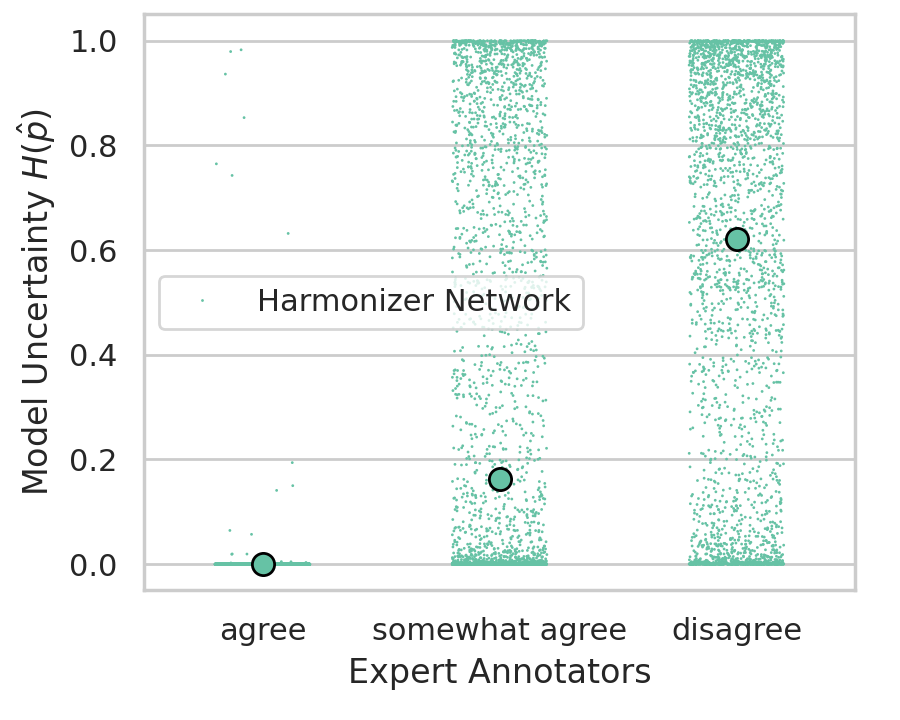}
  \caption{\textbf{Model uncertainty vs. rater agreement.} Pixelwise uncertainty $H(\hat{p})$ grouped by rater consensus: agree, somewhat agree, and disagree. Each dot represents a sampled pixel; large circles denote medians. Uncertainty increases smoothly with annotation ambiguity, indicating that the model's confidence is aligned with inter-rater variability.}
  \label{fig:agreement_uncert}
\end{figure}

Figure~\ref{fig:agreement_uncert} shows the distribution of model uncertainty $H(\hat{p})$ across these categories. Points represent individual pixels (subsampled for clarity), and group medians are marked by large circles. The results reveal a clear monotonic trend: 
\begin{itemize}
    \item In regions where annotators agree, the model expresses low uncertainty, with entropy collapsing near zero.
    \item In regions of partial agreement, uncertainty rises moderately, indicating cautious predictions.
    \item In regions where annotators disagree, model uncertainty is highest, reflecting maximum ambiguity.
\end{itemize}

This behavior demonstrates that the model is not only well-calibrated but also sensitive to human disagreement. Its uncertainty estimates increase precisely in regions where expert interpretation diverges, rather than being uniformly elevated or erratic. This alignment between model and rater uncertainty is critical for practical deployment: it enhances interpretability, facilitates automatic triage of ambiguous regions, and reinforces trust in the model’s outputs.

Moreover, these results complement distributional evaluation metrics such as GED by showing that the model’s uncertainty is not just diverse, but also context-aware. It can differentiate between confident and uncertain predictions in a way that reflects human annotation variability, making it a meaningful tool for decision support and quality control in multi-rater clinical environments.

\section{Per-Rater Calibration: ECE and Brier Score}
\label{sec:calibration}

While Dice and GED quantify overlap-based segmentation accuracy, an equally important property of multi-rater probabilistic models is the \emph{calibration} of their predicted probabilities. A well-calibrated model ensures that predicted confidences, such as $0.3$ or $0.8$, accurately reflect the empirical frequency of foreground pixels. To evaluate this property, we compute the Expected Calibration Error (ECE)~\cite{guo17a} and Brier score~\cite{steyerberg10} for each personalized rater head on the LIDC dataset.

For each rater $A_r$, the personalized head produces a foreground probability map $p_r(x) \in [0,1]$, which is evaluated against that rater’s binary annotation $y_r(x) \in \{0,1\}$. All pixels from all cross-validation folds are pooled together and grouped into $K=10$ equal-width bins $\{B_k\}_{k=1}^K$ spanning the probability range $[0,1]$. For each bin $B_k$, we compute the mean predicted confidence $\text{conf}(B_k)$ and the corresponding empirical foreground frequency $\text{acc}(B_k)$.

The Expected Calibration Error aggregates deviations between confidence and accuracy as
\begin{equation}
\text{ECE}
= \sum_{k=1}^{K} \frac{|B_k|}{N}\,\big|\text{acc}(B_k) - \text{conf}(B_k)\big|,
\end{equation}
where $N$ denotes the total number of evaluated pixels. An ECE value close to zero indicates that the model’s predicted probabilities faithfully correspond to empirical frequencies. Complementarily, the Brier score measures the mean squared error between predicted probabilities and binary labels:
\begin{equation}
\text{Brier}
= \frac{1}{N} \sum_{i=1}^{N} \big(p_i - y_i\big)^2,
\end{equation}
where lower values correspond to more accurate and better-calibrated probabilistic estimates.

\begin{figure}[b]
    \centering
    \includegraphics[width=1\linewidth]{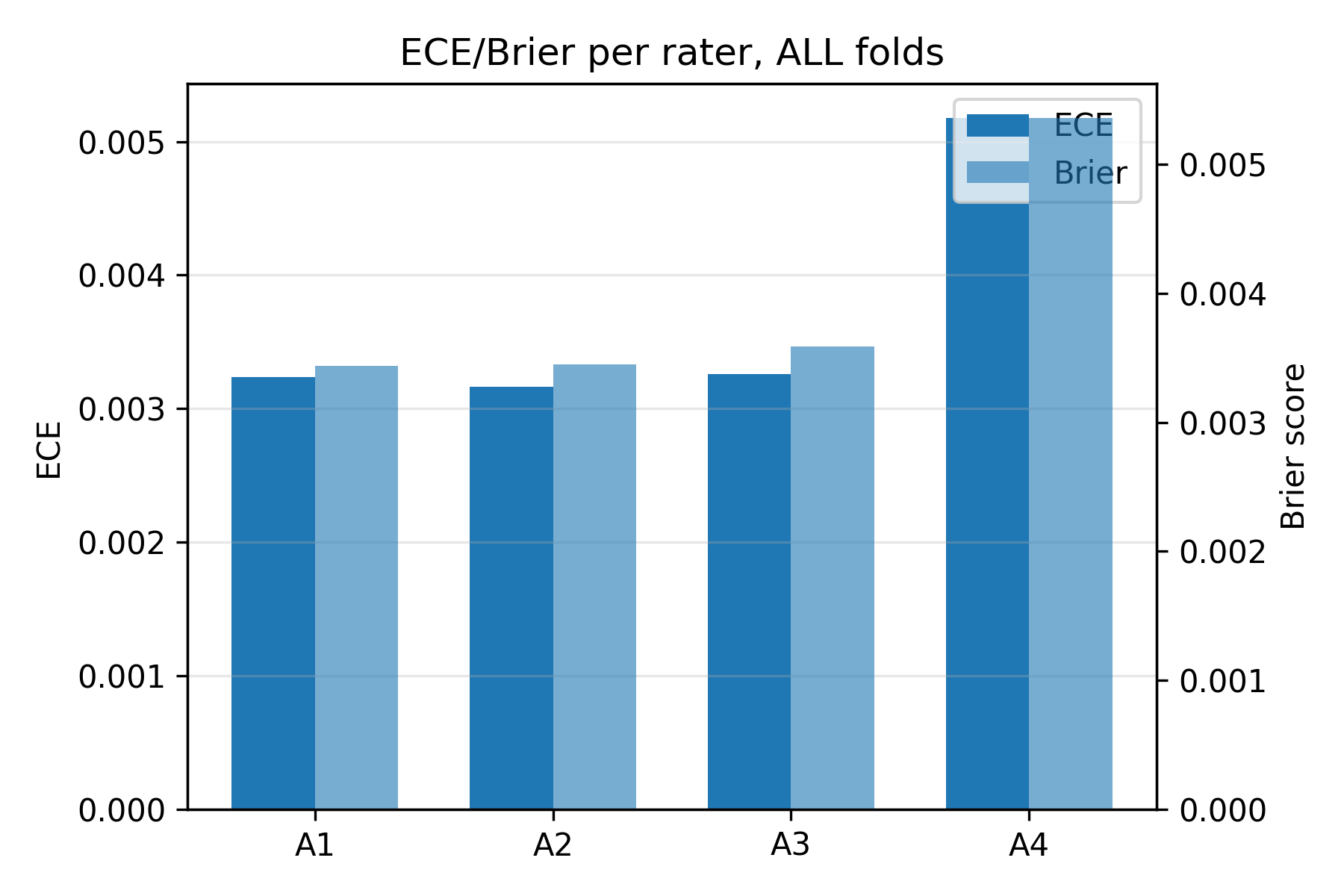}
    \caption{ECE and Brier score per rater, aggregated across all LIDC folds.  
    Bar heights correspond to the numerical values reported in Table~\ref{tab:ece_brier_lidc}.}
    \label{fig:ecebar}
\end{figure}

Table~\ref{tab:ece_brier_lidc} reports per-rater calibration metrics, while Figure~\ref{fig:ecebar} visualizes both ECE and Brier score jointly. All raters achieve extremely low ECE values ($3\!\times\!10^{-3}$–$5\!\times\!10^{-3}$) and Brier scores below $6\!\times\!10^{-3}$, demonstrating strong probabilistic reliability across annotators. Raters A1--A3 exhibit nearly identical calibration behavior, indicating that the model captures their annotation styles consistently. In contrast, Rater A4 shows slightly higher error, which aligns with this rater being the most challenging (e.g., exhibiting sharper boundaries or more aggressive contouring), yet the absolute differences remain very small.

Overall, these results confirm that our Harmonized Personalization framework not only reproduces individual rater styles but also yields well-calibrated and trustworthy probability estimation.

\begin{table}[t]
    \centering
    \caption{Per-rater calibration on LIDC (Personalized). Lower is better for both metrics.}
    \label{tab:ece_brier_lidc}
    \begin{tabular}{lcc}
        \toprule
        Rater & ECE $\downarrow$ & Brier $\downarrow$ \\
        \midrule
        A1 & 0.003238 & 0.003440 \\
        A2 & 0.003166 & 0.003449 \\
        A3 & 0.003260 & 0.003592 \\
        A4 & 0.005175 & 0.005360 \\
        \midrule
        \rowcolor[HTML]{C8FFFD} 
        Mean & 0.00371 & 0.00396 \\
        \bottomrule
    \end{tabular}
\end{table}

\begin{table*}[t]
  \centering
  \caption{Robustness comparison of probabilistic and personalized segmentation methods under synthetic degradations on the LIDC\textendash IDRI dataset. Identical noise configurations are used across models. $K$ denotes the Gaussian blur kernel size. $|\Delta|$ indicates Dice drop relative to the clean image baseline.}
  \label{tab:allinone}
  \resizebox{0.97\textwidth}{!}{
  \begin{tabular}{c|cccccc|cccccc|cccccc}
    \toprule
    \textbf{Method}  & \multicolumn{6}{c|}{\textbf{Gaussian Noise}} & \multicolumn{6}{c|}{\textbf{Gaussian Blur} ($K=115$)} & \multicolumn{6}{c}{\textbf{Gamma Jittering}}\\
    & \multicolumn{2}{c}{$\sigma=0.10$} & \multicolumn{2}{c}{$\sigma=0.15$} & \multicolumn{2}{c|}{$\sigma=0.25$} &  
      \multicolumn{2}{c}{$\sigma=1.0$} & \multicolumn{2}{c}{$\sigma=1.5$} & \multicolumn{2}{c|}{$\sigma=1.75$} &
      \multicolumn{2}{c}{$\gamma=0.75$}& \multicolumn{2}{c}{$\gamma=1.5$} & \multicolumn{2}{c}{$\gamma=1.75$}\\
    \cline{2-19}
     &\textbf{DSC}\bm{$\uparrow$} & \bm{$|\Delta|\downarrow$}& 
      \textbf{DSC}\bm{$\uparrow$}& \bm{$|\Delta|\downarrow$}& 
      \textbf{DSC}\bm{$\uparrow$}& \bm{$|\Delta|\downarrow$}&
      \textbf{DSC}\bm{$\uparrow$}& \bm{$|\Delta|\downarrow$}& 
      \textbf{DSC}\bm{$\uparrow$}& \bm{$|\Delta|\downarrow$}& 
      \textbf{DSC}\bm{$\uparrow$}& \bm{$|\Delta|\downarrow$}&
      \textbf{DSC}\bm{$\uparrow$}& \bm{$|\Delta|\downarrow$}& 
      \textbf{DSC}\bm{$\uparrow$}& \bm{$|\Delta|\downarrow$}& 
      \textbf{DSC}\bm{$\uparrow$}& \bm{$|\Delta|\downarrow$}\\
    \midrule
    Prob.\,U\textendash Net~\cite{kohl2018probabilistic} 
    &85.43 &3.66& 81.76 &7.32& 73.22 &15.87&
     87.75 &1.33& 86.65 &2.43& 85.86 &3.23&
     86.48 &2.60& 79.92 &9.17& 79.03 &10.06\\
    D\textendash Persona~\cite{wu2024diversified}
    &85.74 &3.43& 81.80 &7.37& 71.11 &18.06&
     88.54 &0.63& 87.65 &1.51& 86.91 &2.25&
     87.66 &1.51& 81.85 &7.31& 81.53 &7.64\\
     \rowcolor[HTML]{C8FFFD} 
    \textbf{Harmonizer (ours)}
    &\textbf{89.37} &\textbf{1.44}& \textbf{87.92} &\textbf{2.88}& \textbf{84.27} &\textbf{6.53}&
     \textbf{90.60} &\textbf{0.22}& \textbf{90.25} &\textbf{0.56}& \textbf{89.76} &\textbf{1.04}&
     \textbf{90.14} &\textbf{0.67}& \textbf{86.75} &\textbf{4.06}& \textbf{86.23} &\textbf{4.58}\\
    \bottomrule
  \end{tabular}}
\end{table*}

\section{Noise Effects}
To evaluate robustness under image degradation, we simulate common acquisition artifacts directly on the LIDC\textendash IDRI test set during inference. Three categories of perturbations are introduced: (i) additive Gaussian noise at three intensity levels ($\sigma_1 = 0.1$, $\sigma_2 = 0.15$, $\sigma_3 = 0.25$); (ii) Gaussian blurring with kernel size $115$ and increasing standard deviations ($\sigma_1 = 1.0$, $\sigma_2 = 1.5$, $\sigma_3 = 1.75$); and (iii) photometric distortions via gamma jittering ($\gamma_1 = 0.75$, $\gamma_2 = 1.5$, $\gamma_3 = 1.75$), as summarized in Table~\ref{tab:allinone}. Using identical noise settings across methods ensures a fair comparison. We evaluate the Harmonizer Network against the probabilistic U-Net \cite{kohl2018probabilistic} and the D-Persona~\cite{wu2024diversified} to assess degradation resilience. 
Across all noise regimes, our method exhibits markedly smaller performance drops in both GED and Dice metrics. Under the strongest Gaussian noise ($\sigma{=}0.25$), the Harmonizer retains more than $95\%$ of its clean\textendash data Dice, whereas D\textendash Persona shows a steeper decline. Similar stability is observed under blurring and jitter perturbations, confirming that our harmonization mechanism effectively suppresses low\textendash frequency degradation and preserves structural coherence in the latent space.

\begin{figure*}[h]
\centering
\includegraphics[width=0.93\linewidth]{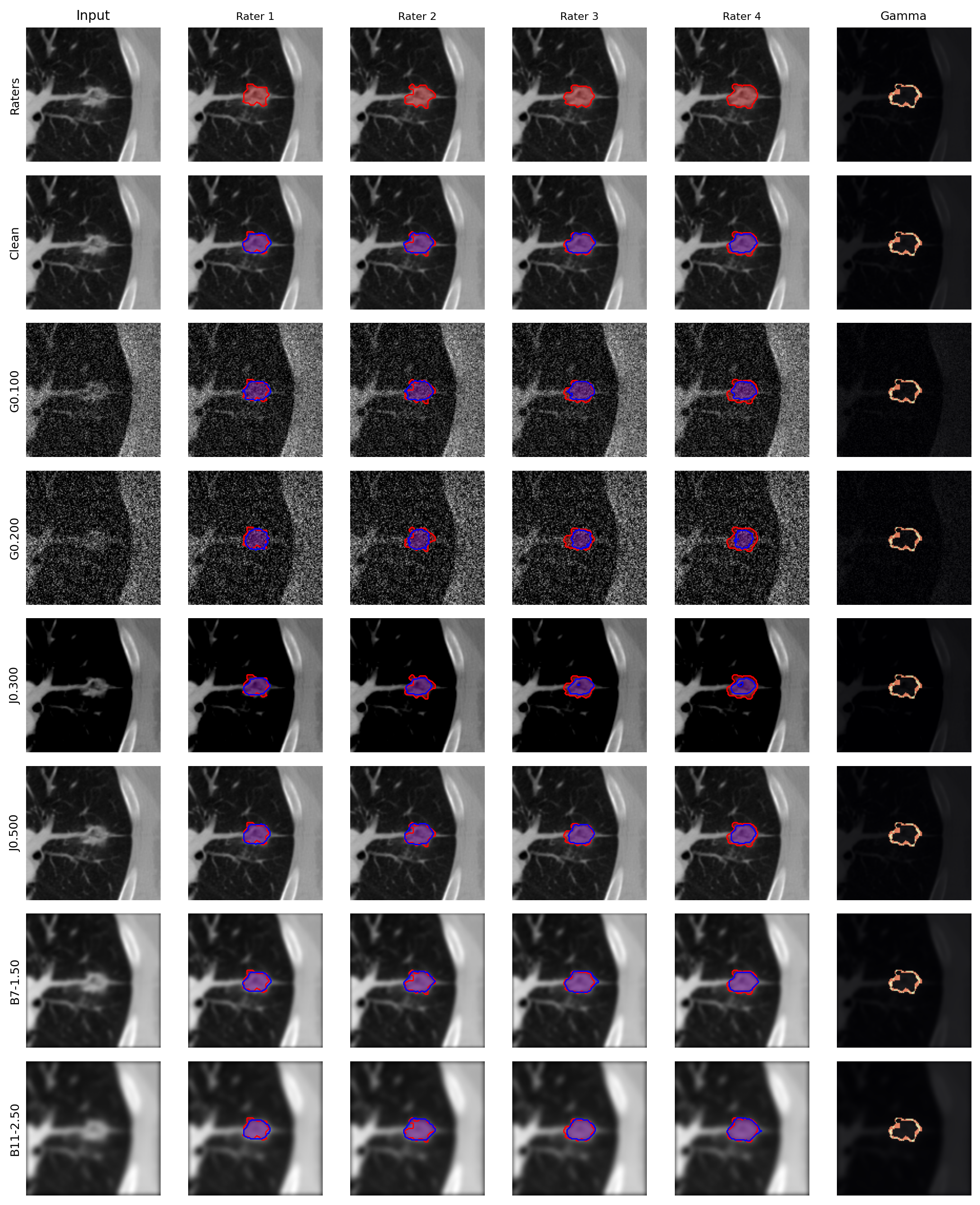}
\caption{
\textbf{Qualitative robustness analysis under noise perturbations.}  
Row labels indicate the applied corruption type and magnitude:
\textbf{G·} denotes Gaussian noise with standard deviation values;  
\textbf{J·} indicates brightness/contrast jitter; and  
\textbf{B·} represents Gaussian blur, where the first and second rows use kernel sizes of 7 and 11, respectively.  
Red contours show ground-truth rater segmentations, blue/purple contours show model predictions, and the rightmost column displays the \textbf{Gamma map}, visualizing inter-rater uncertainty.  
The model preserves structure under all perturbations and degrades meaningfully in regions of high rater ambiguity.
}
\label{fig:noisy}
\end{figure*}

Figure~\ref{fig:noisy} presents a representative example comparing the four expert annotations (red contours), the model prediction on the unperturbed slice (Clean), and predictions obtained under progressively stronger perturbations. Blue/purple contours denote the model’s outputs, while the rightmost column shows the fixed Gamma map that captures inter-rater uncertainty.
Across Gaussian noise conditions (G0.100–G0.200), the model demonstrates remarkable resilience. Even under severe pixel-level corruption, the predicted boundaries remain closely aligned with the raters’ annotations and preserve the overall nodule geometry. The contours do not collapse or drift, instead reflecting a stable consensus shape that balances variations among raters. For instance, when Rater~1 delineates sharper edges while Rater~3 marks smoother boundaries, the model yields an intermediate contour that remains anatomically plausible. This shows that harmonization guides the model toward rater-consistent latent structure rather than superficial textures, effectively suppressing high-frequency noise.

Under intensity jitter (J0.300–J0.500), the model again shows strong invariance. Global brightness and contrast shifts do not affect segmentation alignment; even at higher jitter levels, delineations remain consistent and faithful to the raters. In ambiguous regions, predictions gently interpolate between annotators rather than fluctuating, suggesting that the model prioritizes structural and contextual cues over raw pixel intensities, critical for robustness to scanner variability and acquisition differences.

The Gaussian blur perturbations (B7–B11) reveal the most visually challenging condition. As blur increases, edges become less defined and rater discrepancies more visible: Rater~1 and Rater~4 tend to preserve sharper margins, while Rater~2 and Rater~3 provide smoother contours. The model remains stable across these variations, producing an averaged contour that reflects consensus structure. Under extreme blur (B11–2.50), where the lesion boundary nearly disappears, the prediction contracts slightly toward the core region that all raters agree upon. This behavior is semantically meaningful, rather than failing arbitrarily, the model defaults to the most confident portion of the lesion.

As the Gamma map stays nearly constant across noise types, it serves as a stable reference for uncertainty. Regions that shift under strong blur align with high-uncertainty areas in the Gamma map, while consistent regions remain unaffected. This shows that the model’s sensitivity reflects human disagreement rather than noise artifacts.

\section{Frequency-Domain Visualization}
\begin{figure*}[t]
    \centering
    \begin{subfigure}[t]{0.78\linewidth}
        \centering
        \includegraphics[width=\linewidth]{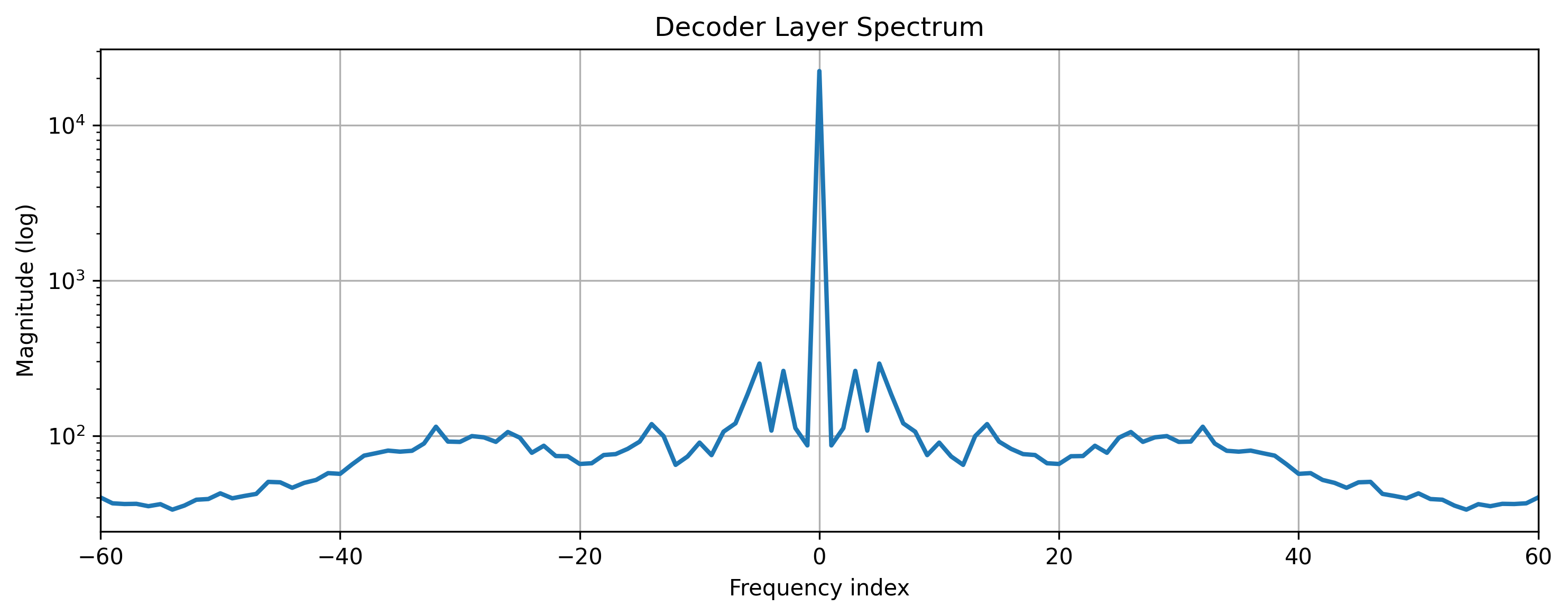}
        \caption{Baseline decoder (without frequency module).}
    \end{subfigure}
    \hfill
    \begin{subfigure}[t]{0.78\linewidth}
        \centering
        \includegraphics[width=\linewidth]{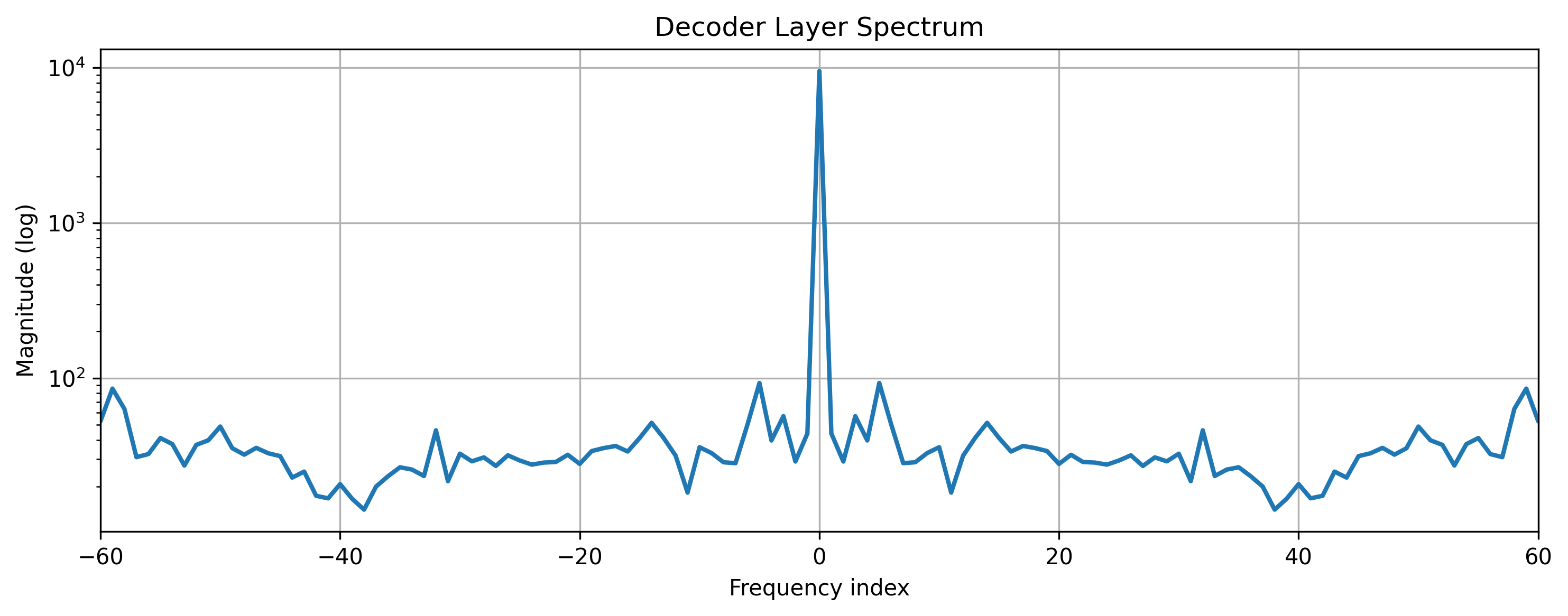}
        \caption{With frequency adapter (ours).}
    \end{subfigure}
    \caption{Spectral response comparison between the baseline decoder and our frequency-calibrated decoder. The plots show the 1D FFT magnitude (log-scale) along the central spatial axis. The adapter increases high-frequency activation while preserving stable low-frequency structure.}
    \label{fig:spectral_comparison}
\end{figure*}

The main purpose of the frequency adapter is to strengthen high-frequency information, enabling the model to better exploit boundary-related and textural cues during personalization. To verify that the module performs this enhancement, we visualize the spectral response of the decoder’s final feature map using a 1D FFT along the central spatial axis, following the procedure in \cite{azad2023laplacian}. Figure~\ref{fig:spectral_comparison} compares the baseline model with our frequency-calibrated one. Without the adapter, the spectrum is dominated by low-frequency components, indicating limited usage of fine details. After applying the adapter, the high-frequency energy increases substantially, while the low-frequency structure remains stable. This confirms that the module effectively amplifies informative high-frequency signals, enriching fine-scale structure in the feature space and enabling sharper, more personalized predictions.

\section{Latent Space Diversity}
\begin{figure*}[t]
    \centering
    \includegraphics[width=0.81\linewidth]{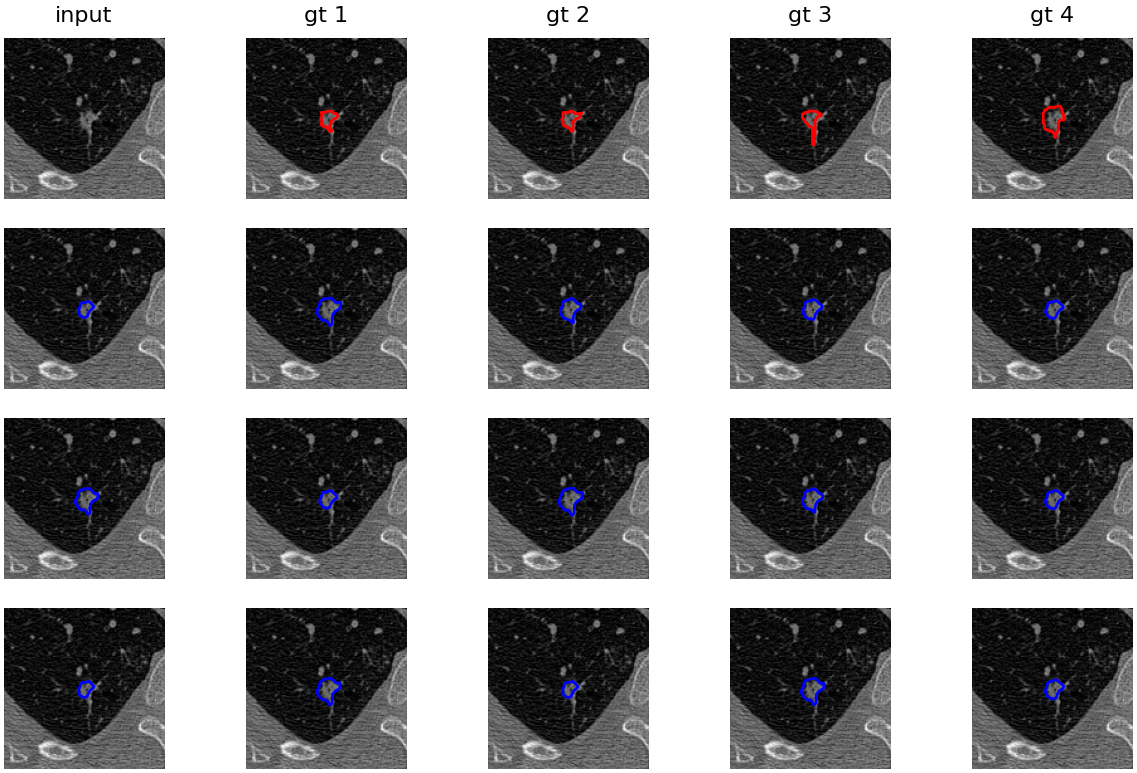}
    \caption{Diverse segmentation hypotheses generated by sampling from the latent space on the LIDC dataset. Each column represents a stochastic sample, illustrating structured diversity along ambiguous boundaries.}
    \label{fig:latentlidc}
\end{figure*}

\begin{figure*}[t]
    \centering
    \includegraphics[width=0.81\linewidth]{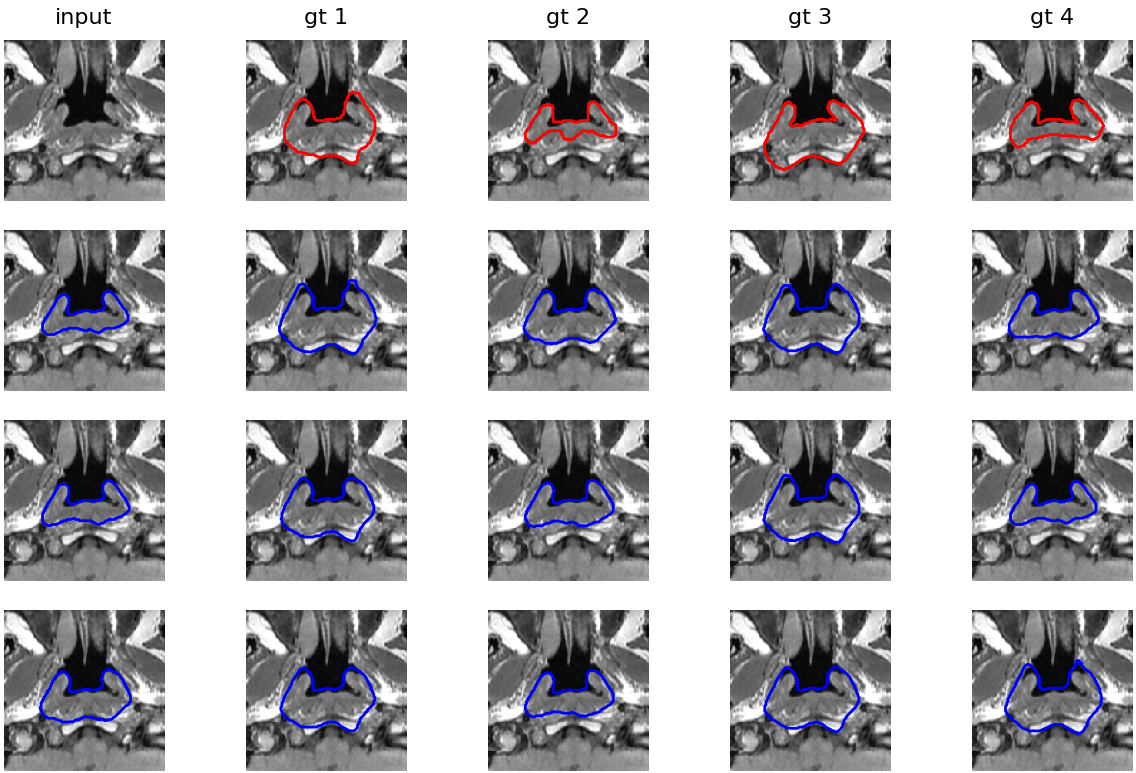}
    \caption{Diverse segmentation hypotheses generated by sampling from the latent space on the NPC-170 dataset. Each column represents a stochastic sample, illustrating structured diversity along ambiguous boundaries.}
    \label{fig:latentnpc}
\end{figure*}

A fundamental property of probabilistic segmentation models is their ability to represent multiple plausible outputs for a given image. 
To assess whether our model’s latent space captures genuine inter-rater diversity rather than random variability, we analyze its generative behavior by sampling multiple predictions from the learned prior distribution. 
Figure~\ref{fig:latentlidc} and Figure~\ref{fig:latentnpc} present representative examples of sampled segmentation masks on the LIDC and NPC-170 datasets, respectively. 
Each row corresponds to a single input image, while the columns depict diverse segmentations generated by different latent codes.

These examples demonstrate that the latent space learned by our method is both structured and diverse. 
Each sample remains anatomically valid while exhibiting subtle yet meaningful variations, particularly along boundary regions where expert annotations typically diverge. 
This diversity indicates that the model does not collapse to a single deterministic solution but instead preserves multiple plausible hypotheses consistent with the empirical annotation distribution. 
The frequency-domain prompts introduced in our framework play a key role in this property by modulating harmonized features within a rater-aware spectral space. 
Consequently, the model maintains controlled variability concentrated in clinically ambiguous regions rather than distributing it arbitrarily across the image.

\begin{figure}[thb]
    \centering
    \includegraphics[width=1\linewidth]{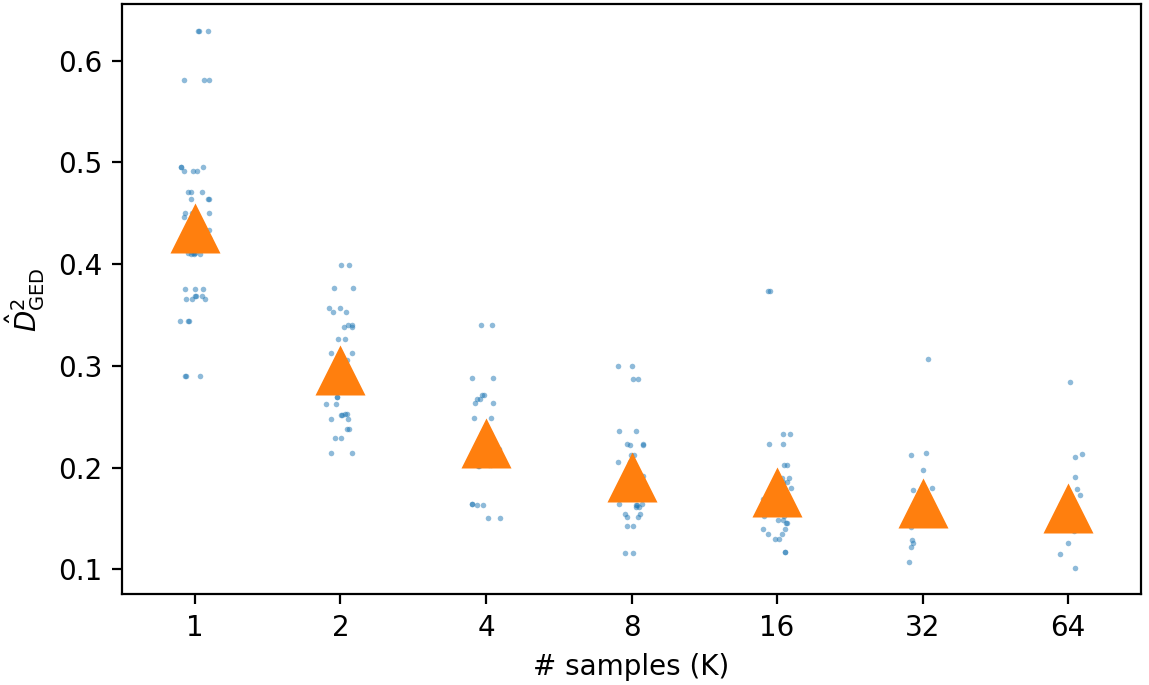}
    \caption{
    Impact of the proposed GED loss on distributional coverage. 
    Plot shows GED as a function of the number of generated samples \(K\) (lower is better). 
    Models trained with the GED loss achieve consistently lower discrepancy and reach convergence with fewer samples, indicating more efficient and faithful coverage of the underlying annotation distribution.
    }
    \label{fig:gednpc}
\end{figure}

To quantify latent diversity, we also evaluate the distributional coverage using GED as a function of the number of generated samples on the NPC-170 dataset (Figure~\ref{fig:gednpc}). 
We observe a consistent reduction in GED, with a simultaneous increase in soft-Dice as the number of samples grows, confirming that the model efficiently spans the annotation manifold. 
Importantly, the improvements saturate beyond approximately 30 samples, suggesting that the latent space is compact yet expressive enough to capture the full range of expert annotations without redundancy. 
These findings provide strong evidence that the learned latent representation encodes realistic structural uncertainty and enables faithful multi-rater modeling through controlled stochastic sampling.

\section{Extended Results on Kvasir Dataset}

We further evaluate our approach on the noisy medical image segmentation dataset Kvasir~\cite{jha2019kvasir}, which presents varying annotation challenges. As shown in Table~\ref{tab:performance_comparison}, our method achieves the best overall performance on the Kvasir dataset.

\subsection*{Dataset and Experimental Setup}
The Kvasir-SEG dataset~\cite{jha2019kvasir} contains 1,000 gastrointestinal endoscopy images, each paired with an expert-annotated polyp segmentation mask. The images exhibit substantial variability in polyp size, morphology, and visual appearance, and are acquired under heterogeneous illumination conditions. Following the noise-injection protocol of~\cite{wang2025noisy}, we simulate annotation noise and inter-rater disagreement by generating three additional mask variants for each training image: $\text{S}_\text{R}$ (Simulated Random noise, $\sigma=0.2$), $\text{S}_\text{E}$ (Simulated Extreme noise, $\sigma=0.8$), and $\text{S}_\text{DE}$ (Simulated Diverse Expert noise). This yields four masks per training sample. The test set uses only clean expert annotations to ensure unbiased evaluation. In our experiments, we follow~\cite{wang2025noisy} and resize images to $256 \times 256$ and normalize them to $[0,1]$.

\begin{table}[htbp]
\centering
\fontsize{9}{11}\selectfont
\setlength{\tabcolsep}{4pt}
\caption{Performance comparison on Kvasir dataset}
\label{tab:performance_comparison}
\begin{tabular}{@{}lccc@{}}
\toprule
\multirow{2}{*}{\textbf{Method}} & \multicolumn{3}{c}{\textbf{Kvasir dataset}} \\
\cmidrule(l){2-4}
 & \textbf{$S_R$} & \textbf{$S_E$} & \textbf{$S_{DE}$} \\
\midrule
RCE Loss~\cite{wang2019symmetric} & 73.51±1.58 & 73.68±1.57 & 66.17±1.97 \\
RMD~\cite{fang2023reliable} & 68.34±2.18 & 71.57±1.09 & 66.90±1.75 \\
ADELE~\cite{liu2022adaptive} & 60.97±14.78 & 67.10±11.42 & 60.62±13.04 \\
CDR~\cite{xia2020robust} & 67.87±3.51 & 70.58±1.65 & 63.51±1.67\\
Co-Teaching~\cite{han2018co} & 74.26±1.71 & 75.57±1.12 & 74.03±1.48 \\
IDMPS~\cite{zhao2024ultrasound} & 77.52±1.21 & 74.16±0.81 & 69.47±0.81 \\
JoCoR~\cite{wei2020combating} & 67.98±3.23 & 71.43±1.37 & 65.62±1.94 \\
SP-Guide~\cite{li2021superpixel} & 69.24±2.09 & 62.97±1.65 & 61.74±1.56 \\
GSD-Net~\cite{wang2025noisy} & 80.04±0.42 & 79.39±0.33 & \textbf{79.97±0.53} \\
D-Persona~\cite{wu2024diversified} & 84.69±0.19 & 81.77±0.16 & 78.93±0.18 \\
\rowcolor[HTML]{C8FFFD} 
Harmonizer Network &\textbf{85.13±0.17} &\textbf{82.96±0.15} &{78.89±0.18} \\
\bottomrule
\end{tabular}
\end{table}

\subsection*{Results}
\begin{figure*}[t]
\centering
\resizebox{0.90\linewidth}{!}{
\begin{tabular}{cccc}
\textbf{Input} & \textbf{GT Mask} & \textbf{Prediction Heatmap} & \textbf{Prediction} \\
\\[-0.5em]
\includegraphics[width=0.22\linewidth]{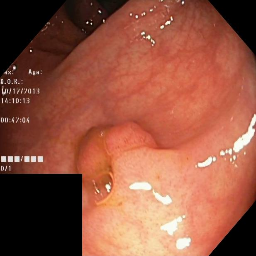} &
\includegraphics[width=0.22\linewidth]{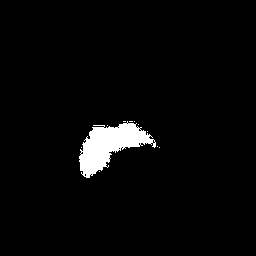} &
\includegraphics[width=0.22\linewidth]{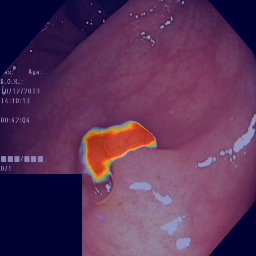} &
\includegraphics[width=0.22\linewidth]{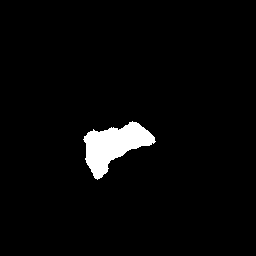} \\
\\[0.5em]
\includegraphics[width=0.22\linewidth]{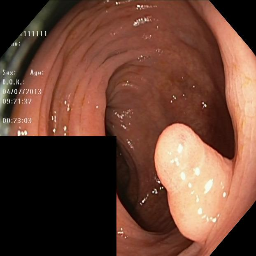} &
\includegraphics[width=0.22\linewidth]{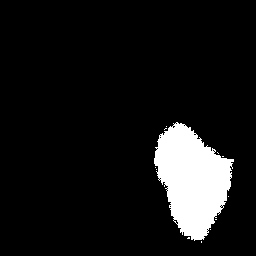} &
\includegraphics[width=0.22\linewidth]{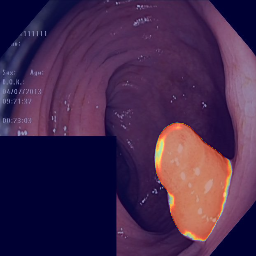} &
\includegraphics[width=0.22\linewidth]{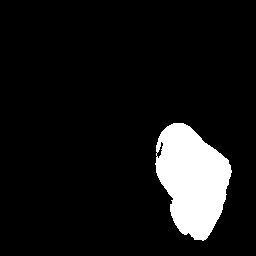} \\
\\[0.5em]
\includegraphics[width=0.22\linewidth]{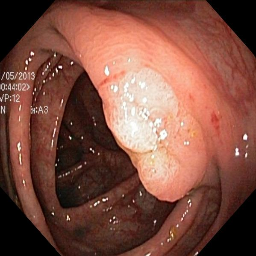} &
\includegraphics[width=0.22\linewidth]{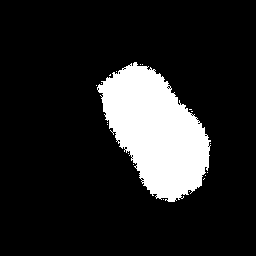} &
\includegraphics[width=0.22\linewidth]{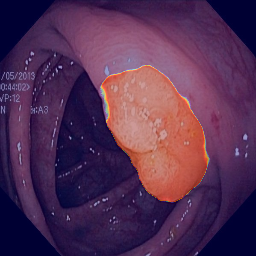} &
\includegraphics[width=0.22\linewidth]{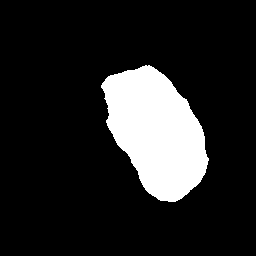} \\
\end{tabular}
}
\caption{Visual comparison of segmentation results on the Kvasir dataset, showing input images, ground truth masks, prediction heatmaps, and final predictions for three representative samples.}
\label{fig:visual_comparison}
\end{figure*}

Table~\ref{tab:performance_comparison} presents a comprehensive comparison of our Harmonizer Network against baseline methods on the Kvasir dataset across three noise conditions. Our Harmonizer Network achieves the highest overall performance across all noise conditions, with Dice scores of 85.13\%, 82.96\%, and 78.89\% for $\text{S}_\text{R}$, $\text{S}_\text{E}$, and $\text{S}_\text{DE}$, respectively. The probabilistic method D-Persona~\cite{wu2024diversified} ranks second with scores of 84.69\%, 81.77\%, and 78.93\% for $\text{S}_\text{R}$, $\text{S}_\text{E}$, and $\text{S}_\text{DE}$, while the GSD-Net~\cite{wang2025noisy} achieves third place with consistent performance around 80\% across all conditions. Compared to D-Persona, our method shows modest but consistent improvements of +0.44\% on $\text{S}_\text{R}$, +1.19\% on $\text{S}_\text{E}$, and -0.04\% on $\text{S}_\text{DE}$. While D-Persona slightly outperforms our method on diverse expert noise, our approach shows superior robustness to extreme noise conditions, suggesting better handling of severe annotation errors through the harmonization strategy. Our method outperforms GSD-Net by +5.09\% and +3.57\% on random and extreme noise conditions, respectively, while trailing by -1.08\% on diverse expert noise. The larger gains on random and extreme noise indicate that our approach more effectively filters noisy labels compared to traditional co-teaching strategies. Co-Teaching~\cite{han2018co} achieves 74.26--75.57\%, while JoCoR~\cite{wei2020combating} reaches only 65.62--71.43\%, both of which fall significantly behind our method. Loss-based approaches show mixed results, with RCE Loss~\cite{wang2019symmetric} achieving moderate performance (66.17--73.68\%), while RMD~\cite{fang2023reliable} and CDR~\cite{xia2020robust} perform similarly (63.51--71.57\%). These methods lag behind our approach by 11--19 percentage points, highlighting the limitations of purely loss-based noise handling without explicit multi-annotator modeling. IDMPS~\cite{zhao2024ultrasound} shows strong performance on $\text{S}_\text{R}$ (77.52\%) but degrades on $\text{S}_\text{E}$ (74.16\%) and $\text{S}_\text{DE}$ (69.47\%). SP-Guide~\cite{li2021superpixel}, performs poorly on extreme noise (62.97\%), suggesting that hand-crafted priors struggle with severe annotation errors. Our method demonstrates superior stability with low standard deviations (±0.15--0.18) compared to most baselines, particularly ADELE (±11--15\%), CDR (±1.65--3.51\%), and JoCoR (±1.37--3.23\%), indicating that our harmonization approach provides more consistent training dynamics across different noise realizations. Additionally, we provide visual comparisons between our model's outputs and other baseline methods in Figure~\ref{fig:visual_comparison}, qualitatively illustrating the superior segmentation quality and boundary accuracy achieved by our approach across different noise conditions.

\section{Model Complexity and Computational Efficiency}

Although the proposed framework integrates both harmonization and personalization mechanisms, it remains lightweight and computationally efficient. 
The full model contains 30.31\,M parameters (baseline 30.11\,M, $\mathcal{H}_\phi^{(n)}$ 0.14\,M, and $\mathcal{H}_\phi^{(p)}$ 0.07\,M), including both harmonization and personalization modules, and requires around 0.42\,s per forward pass during inference. 
We estimate the computational complexity using multiply–accumulate operations (MACs) on $128 \times 128$ input images. The full Harmonizer Network requires approximately 2.4 GFLOPs per forward pass. This marginal cost demonstrates that the added modules perform lightweight modulation rather than heavy computational processing. 
During inference, the model runs at approximately 2.3 frames per second on a single NVIDIA RTX 3090 GPU, which includes latent sampling and personalization. Training employs mixed-precision arithmetic, further reducing memory consumption and accelerating convergence. In fact our design remains significantly efficient while offering higher fidelity and interpretability. The harmonizer’s shared weights across layers and the low-dimensional prompt tokens ensure scalability without exponential growth in computational cost. This makes the model suitable for deployment in real-world medical imaging settings, where computational resources and latency constraints are critical considerations. The lightweight design underscores that harmonized probabilistic modeling, along with frequency-domain personalization, can be achieved efficiently without sacrificing performance or uncertainty quality.

\begin{figure*}[tbh]
    \centering
    \includegraphics[width=0.95\linewidth]{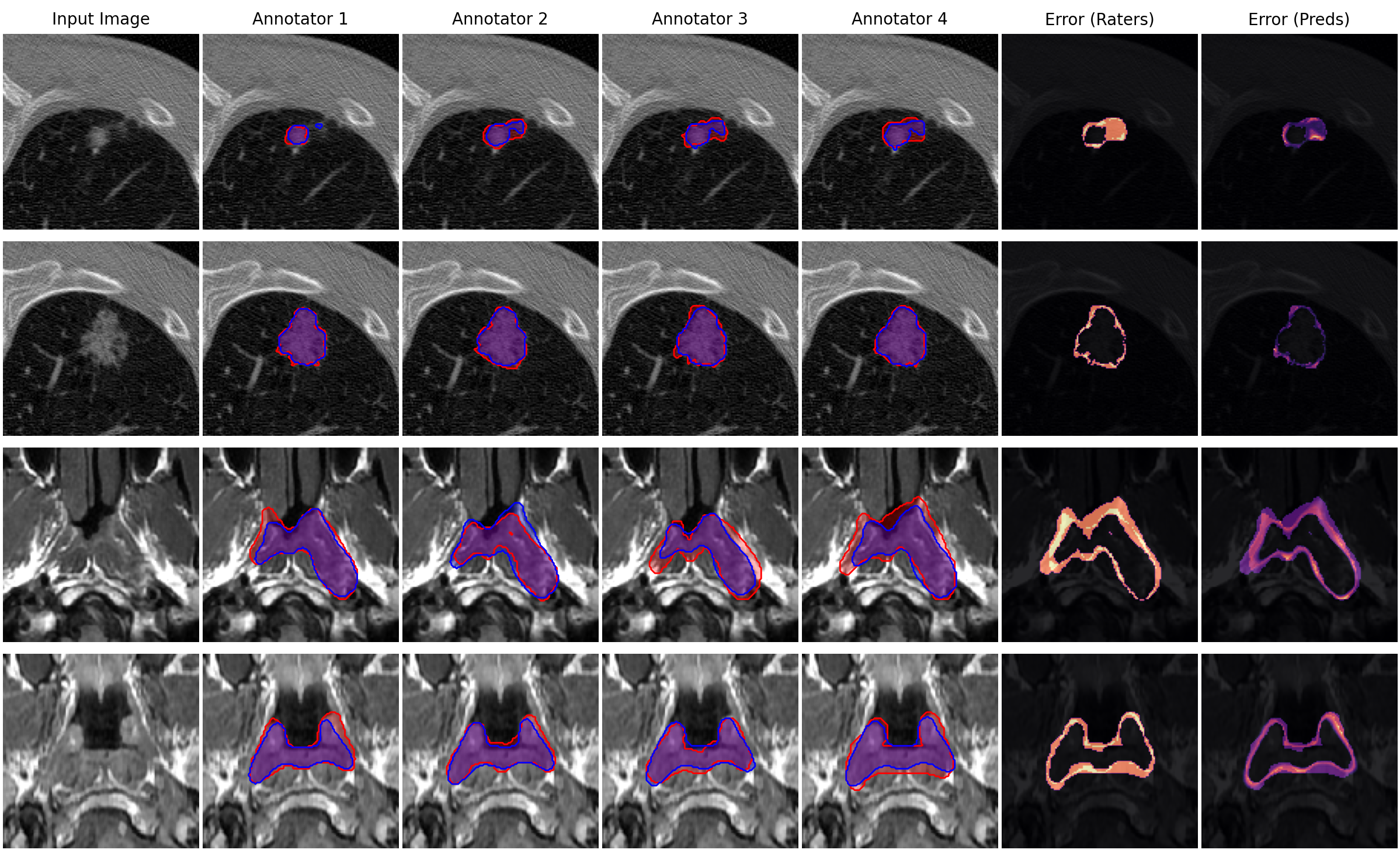}
    \caption{Visualization results on the LIDC (first-second rows) and NPC-170 (third-fourth rows) dataset with multi-rater annotations and the corresponding error map. The red boundary indicates the ground truth, while the blue boundary represents the predicted mask.}
    \label{fig:resultslidc}
\end{figure*}

\subsection{Two-stage training/inference pseudo-code}

This section summarizes the overall optimization and inference pipeline of the proposed framework. The pseudo-code is organized into two consecutive training phases to clearly separate the learning of data-level harmonization from the learning of rater-specific personalization. In the first phase, the encoder $E$, decoder $D$, and Noise Harmonizer $H$ are jointly optimized so that the network can reconstruct segmentations while learning artifact-reduced and stable feature representations. This stage focuses on capturing the underlying anatomical content and the global ambiguity of the task without yet specializing to individual annotators. In the second phase, these backbone components are frozen and only the Personalizer $P$ is trained. This design ensures that rater-specific adaptation is learned on top of already harmonized features, preventing scanner noise or acquisition bias from being mixed with annotator-dependent variation. During inference, the same harmonized representation is first extracted, and then the personalizer conditions the latent code for a selected rater to generate the corresponding individualized segmentation.

\footnotesize

\noindent\textbf{Pseudocode}
\vspace{2mm}
\hrule
\vspace{1mm}
\noindent\textbf{Phase~1 (train $E,D,H$):}

$f=E(x)$, $z\!\sim\!q(z|x,y)$, $\hat f=H(f)$, $\hat y=D(\hat f,z)$;

optimize $L_{\text{ELBO}}+\lambda_{\text{GED}}L_{\text{GED}}$.
Freeze $E,D,H$.

\noindent\textbf{Phase~2 (train $P$ only):} 

$\hat f=H(E(x))$;
for each rater $r$: $z\!\sim\!p(z|x)$,
$z'=\mathcal{P}(z,\hat f,r)$,
$\hat y^r=D(\hat f,z')$;   

optimize $\sum_{r=1}^n\mathbb{E}_{z\sim p(z|x)}[L_{\text{seg}}(\hat y^r,y^r)]$ (Eq.~6).

\textbf{[$z$: prior latent, $z'$: rater-adaptive latent]}

\noindent\textbf{Inference:}
$\hat f=H(E(x))$,
$z\!\sim\!p(z|x)$,
$z'=\mathcal{P}(z,\hat f,r)$,
$\hat y^r=D(\hat f,z')$.
\normalsize
\hrule
\vspace{2mm}

The pseudo-code highlights the functional role of each module in the proposed training strategy. In Phase~1, the latent variable is sampled from the posterior distribution $q(z|x,y)$, allowing the model to learn a structured representation of segmentation uncertainty while the harmonizer refines encoder features into a more stable and artifact-suppressed form. The joint objective combines the ELBO loss with the GED term so that the model not only reconstructs plausible outputs but also captures distributional diversity across annotations. Once this representation is learned, $E$, $D$, and $H$ are frozen. In Phase~2, the prior latent $z\sim p(z|x)$ is adapted by the personalizer into a rater-aware latent code $z'$, which injects annotator-specific information without altering the shared anatomical backbone. This makes the second stage focused and stable, since only the personalization branch is optimized. At inference time, the framework follows the same decomposition: harmonized features are first extracted, then the latent sample is personalized for a chosen rater, and finally the decoder generates the corresponding segmentation. In this way, the model preserves both common anatomical structure and individualized annotation style in a controlled manner.

\section{Additional Highlights}
\subsection*{Importance of Each Module}
The proposed framework integrates two purpose-built modules, the Noise Harmonizer and the Frequency-Prompt Personalizer, each addressing a fundamentally different challenge in the multi-rater segmentation setting. Importantly, both components are designed not to artificially inflate accuracy on clean datasets, but to explicitly resolve the core sources of disagreement and inconsistency inherent in multi-annotator data.

The Noise Harmonizer regulates high-frequency corruption and scanner-induced variability, ensuring that the latent space remains stable even under significant perturbations. As demonstrated in the robustness experiments on Gaussian noise, blur, and jitter (Section 7), this module substantially improves resilience to degradation. Under the strongest corruption level ($\sigma = 0.25$), the Harmonizer retains over 95\% of its clean-data Dice, outperforming D-Persona and Probabilistic U-Net by large margins. Conversely, removing this module on clean LIDC/NPC-170 samples yields only marginal performance changes, consistent with its intended role as a stabilizer rather than a direct accuracy booster.

The Frequency-Prompt Personalizer provides rater-specific adaptation by learning localized spectral modulations that reproduce annotators’ stylistic tendencies without overwhelming the shared anatomical manifold. Section 5 shows that its effect is spatially precise, primarily enhancing high-frequency energy around ambiguous boundaries. Compared with the projection heads of D-Persona, the proposed personalizer consistently yields lower GED, higher soft-Dice, and stronger rater-aligned calibration (Sections~2 and~6). Across NPC-170 and LIDC, these improvements translate into observable gains in segmentation fidelity and more coherent personalized outputs.

Together, the two modules offer complementary gains: one stabilizes the latent representation under real-world noise, while the other enriches stylistic personalization.

\subsection*{Comparison with Consensus-Based Methods}
Consensus-based or label-fusion methods operate by collapsing multiple annotations into a single target, often through majority voting, STAPLE, or random sampling. As highlighted in prior work (e.g., D-Persona), such approaches inevitably discard meaningful rater-specific variations and tend to underperform in settings with genuine annotation disagreement. Our empirical findings are consistent with this trend~\cite{wu2024diversified}.

\begin{figure*}[thb]
    \centering
    \begin{minipage}[t]{0.48\linewidth}
        \includegraphics[width=\linewidth]{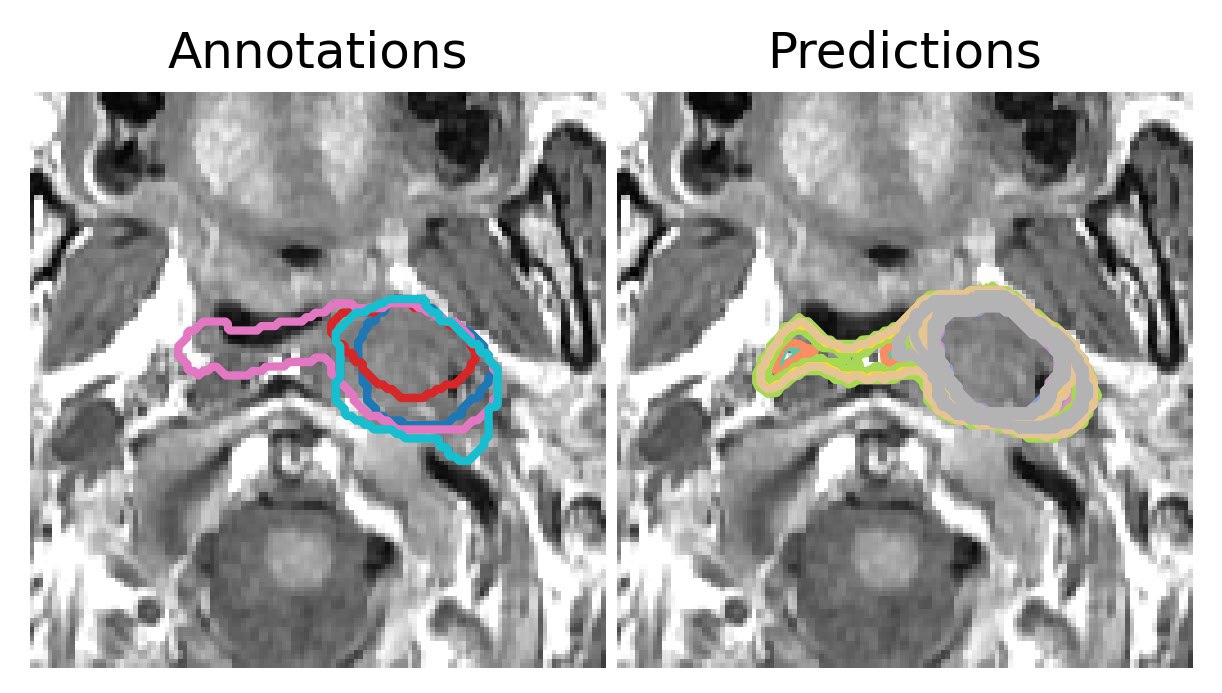}
    \end{minipage}
    \hfill
    \begin{minipage}[t]{0.48\linewidth}
        \includegraphics[width=\linewidth]{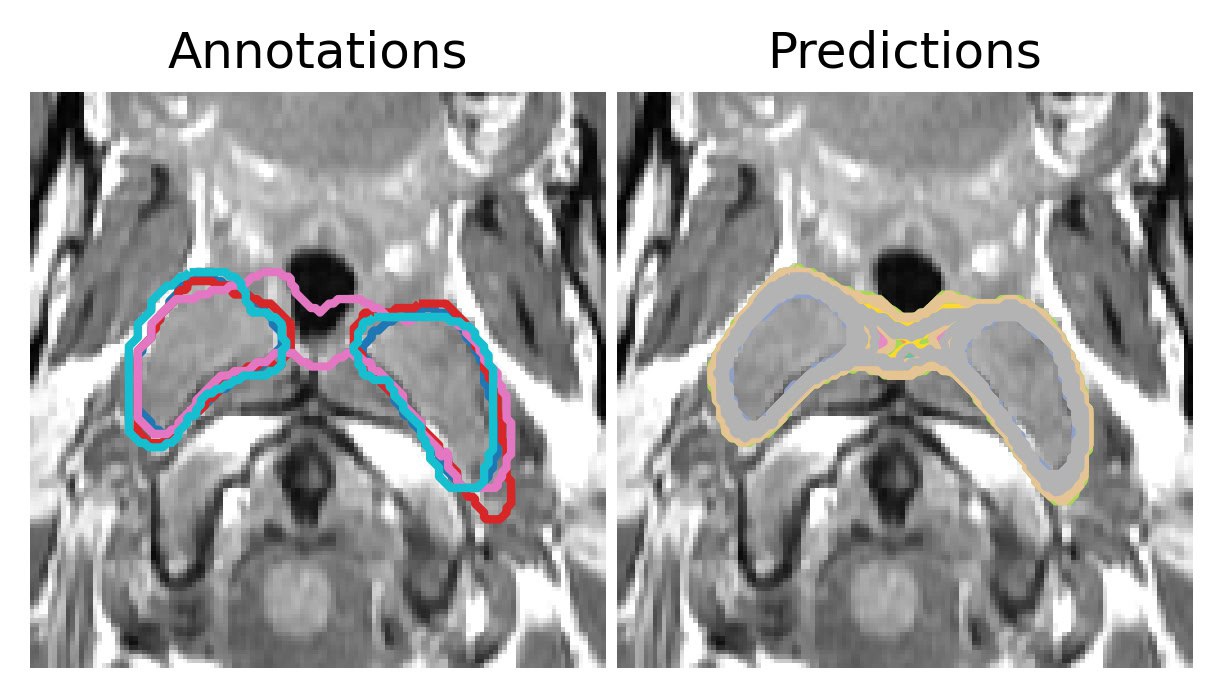}
    \end{minipage}
    \caption{
    \textbf{Representative failure cases.} 
    \textit{Left:} One annotator marks a much larger region than others; the model detects similar patterns but is constrained by rater-specific regularization, yielding a truncated prediction.  
    \textit{Right:} Two symmetric structures lead to inconsistent rater choices. The model captures both hypotheses but cannot fully disambiguate them.  
    Residual errors thus stem from annotation inconsistency rather than model instability.
    }
    
    \label{fig:limitations}
\end{figure*}

\subsection*{Additional Visualization}
To further demonstrate the qualitative behavior of our model, we provide additional visualizations of segmentation results on both the LIDC-IDRI and NPC-170 datasets in Figure~\ref{fig:resultslidc}. These visual samples highlight how the model effectively adapts to heterogeneous annotator styles while preserving structural consistency across multiple raters. In both datasets, we observe that the proposed Harmonizer Network produces highly coherent boundaries that align well with the ground-truth masks, even when annotations diverge. The personalized predictions successfully replicate subtle stylistic tendencies of individual annotators, such as sharper contour delineation or smoother regional filling, demonstrating the model’s ability to internalize each rater’s segmentation characteristics.

The accompanying error maps visualize regions of disagreement between predicted and annotated masks. Red boundaries indicate ground-truth annotations, whereas blue lines denote predicted segmentations. Areas with overlapping red-blue contours correspond to high agreement, while isolated colored edges represent zones of residual ambiguity. These differences often coincide with clinically uncertain regions, low-contrast lesion edges, weak tissue boundaries, or ambiguous nodular margins where experts themselves disagree. The proposed model allocates uncertainty precisely in such areas, confirming that its confidence modulation and personalized adaptation reflect genuine inter-rater ambiguity rather than random noise. These results further demonstrate that the method maintains both personalization and structural consistency across multiple annotation distributions, effectively capturing the underlying anatomical manifold while respecting individual rater styles.

\section{Acquisition-Domain Shifts}
To directly address acquisition domain shift, we conducted a new experiment using the manufacturer metadata provided in the LIDC-IDRI dataset.
We partitioned the LIDC-IDRI dataset based on scanner manufacturer to create a realistic cross-domain evaluation scenario:
\textit{Training/Validation Set}: All samples from GE MEDICAL SYSTEMS, TOSHIBA, and Philips manufacturers
\textit{Test Set}: All samples acquired exclusively from SIEMENS scanners

\begin{table}[h]
\caption{Personalized segmentation performance on LIDC--IDRI under different train/test scaners configuration.}
\centering
\scriptsize
\setlength{\tabcolsep}{2pt}
\renewcommand{\arraystretch}{0.9}
\begin{tabular}{l|cc|cc}
\toprule
Method
&\multicolumn{2}{c|}{\shortstack{Train/Test\\All $\rightarrow$ All}}
&\multicolumn{2}{c}{\shortstack{Train/Test\\All (except Siemens) $\rightarrow$ Siemens}}\\
\cline{2-5}
& DSC $\uparrow$ & $|\Delta| \downarrow$
& DSC $\uparrow$ & $|\Delta| \downarrow$ \\
\midrule
D-Persona
& 89.17 & -
& 83.02 & 6.15 \\
Harmonizer Network
& \textbf{90.78} & -
& \textbf{85.30} & \textbf{5.48} \\
\bottomrule
\end{tabular}
\label{tab_scanner}
\end{table}

Table \ref{tab_scanner} demonstrates that our method exhibits superior robustness to real acquisition-domain shifts compared to D-persona. This performance gap along with  synthetic perturbation experiments provides direct empirical evidence that our Noise Harmonizer is not merely acting as a general feature regularizer, but is specifically modeling and mitigating artifacts. Moreover, the Noise Harmonizer uses weight sharing across all decoder layers, which enforces consistent artifact-suppression behavior across multiple spatial resolutions. Acquisition artifacts such as intensity drift, reconstruction noise, and motion blur tend to exhibit scale-consistent patterns, whereas rater variability is typically localized around object boundaries and varies spatially across cases. By learning shared, scale-consistent modulation parameters $(\gamma_l, \beta_l)$, the harmonizer is biased toward capturing global, low-to-mid frequency perturbations that are characteristic of acquisition artifacts. The Personalization Module trained in Phase~2 then operates on these pre-harmonized features and explicitly models high-frequency, boundary-level variations associated with individual rater styles.

\section{Limitations}
Despite its overall robustness, the framework exhibits several limitations that are closely tied to the complexity and inconsistency of multi-rater annotation behavior. As illustrated in Figure~\ref{fig:limitations}, we identify two representative failure scenarios.  

First, in the left sample, one annotator delineates a noticeably larger region than all others, often because they perceived subtle texture cues that other experts ignored. In these cases, our model partially detects those underlying patterns as well, hence several personalized heads show activations in the same extended region. However, the frequency-prompt personalization module simultaneously constrains each prediction to match the stylistic tendencies of the corresponding annotator. This dual mechanism, pattern detection versus stylistic constraint, can create a bottleneck: the model recognizes a plausible region but is restricted from fully expressing it if it contradicts the annotator’s typical behavior. As a result, the personalized mask may appear truncated relative to the annotator’s unusually large annotation, producing a small drop in per-rater quantitative metrics.

Second, we observe cases where the anatomical structure exhibits two overlapping or symmetric patterns, both of which could be interpreted as valid targets. Some annotators segment both regions, while others label only one. This mixture of plausible but inconsistent interpretations makes it inherently difficult for the model to disambiguate the “intended” target for each annotator. While the harmonized latent space captures both spatial hypotheses, the personalized decoding sometimes struggles to resolve the ambiguity when annotator styles diverge only subtly. This further reflects the dependency on annotation consistency: when multiple interpretations are equally defensible, perfect personalization becomes inherently ill-posed.


\end{document}